\documentclass[journal]{IEEEtran}
\ifCLASSINFOpdf
\else
\fi
\usepackage{framed,multirow}

\usepackage{amssymb}
\usepackage{amsmath}
\usepackage{latexsym}
\usepackage{booktabs, caption, makecell}

\usepackage{bm}
\usepackage[english]{babel}
\usepackage{url}
\usepackage{xcolor}
\usepackage{graphics}
\usepackage{graphicx}
\usepackage{subfigure}
\usepackage{caption}
\usepackage{algorithm,algorithmic}
\usepackage[bottom,flushmargin,hang,multiple]{footmisc}

\begin{document}
%
\title{Deep Ordinal Hashing with Spatial Attention}
%
%
%

\author{Lu~Jin,
        Xiangbo~Shu,
        Kai~Li,
        Zechao~Li,
        Guo-Jun~Qi,
        Jinhui~Tang,~\IEEEmembership{Senior Member,~IEEE}
\IEEEcompsocitemizethanks{\IEEEcompsocthanksitem L. Jin, X. Shu, Z. Li and J. Tang are with the School of Computer Science and Engineering, Nanjing University of Science and Technology, Nanjing, 210094, China, (email:lujin505@gmail.com, shuxb@njust.edu.cn, zechao.li@njust.edu.cn and jinhuitang@njust.edu.cn). J. Tang is the corresponding author.
\IEEEcompsocthanksitem K. Li is with Facebook, Menlo Park, CA, 94025, USA, (email:kailee88@fb.com).
\IEEEcompsocthanksitem G. Qi are with the Department of Computer Science, University of Central Florida, Orlando, FL, 32816, USA, (email:kaili@eecs.ucf.edu and guojun.qi@ucf.edu).}}

%
%

\markboth{Submission for IEEE Transactions on Image Processing}%
{Shell \MakeLowercase{\textit{et al.}}: Bare Demo of IEEEtran.cls for IEEE Journals}
%



\maketitle

\begin{abstract}
Hashing has attracted increasing research attentions in recent years due to its high efficiency of computation and storage in image retrieval. Recent works have demonstrated the superiority of simultaneous feature representations and hash functions learning with deep neural networks. However, most existing deep hashing methods directly learn the hash functions by encoding the global semantic information, while ignoring the local spatial information of images. The loss of local spatial structure makes the performance bottleneck of hash functions, therefore limiting its application for accurate similarity retrieval. In this work, we propose a novel Deep Ordinal Hashing (DOH) method, which learns ordinal representations by leveraging the ranking structure of feature space from both local and global views. In particular, to effectively build the ranking structure, we propose to learn the rank correlation space by exploiting the local spatial information from Fully Convolutional Network (FCN) and the global semantic information from the Convolutional Neural Network (CNN) simultaneously. More specifically, an effective spatial attention model is designed to capture the local spatial information by selectively learning well-specified locations closely related to target objects. In such hashing framework, the local spatial and global semantic nature of images are captured in an end-to-end ranking-to-hashing manner.
Experimental results conducted on three widely-used datasets demonstrate that the proposed DOH method significantly outperforms the state-of-the-art hashing methods.
\end{abstract}


%
\IEEEpeerreviewmaketitle

\section{Introduction}
%
%
%
%
\IEEEPARstart{R}{ecently}, large-scale image retrieval has attracted wide attentions in the field of computer vision due to the rapid advancement of information techniques. With the explosive growth of multimedia data including images and videos, hashing has received a great deal of attentions in large-scale visual retrieval for its capability in storage and computation efficiency \cite{gionis1999similarity,datar2004locality,jain2008fast,broder2000min,gong2013iterative,kulis2009kernelized,lin2013general,salakhutdinov2009semantic,norouzi2012hamming}. Hashing is to construct a set of hash functions by projecting the high dimensional data in the visual space into compact binary codes in the Hamming space. Due to the amazing performance of deep features in computer vision tasks, such as image classification \cite{krizhevsky2012imagenet,ciregan2012multi,oquab2014learning}, image captioning \cite{chen2016sca,you2016image,yang2016review} and image retrieval \cite{liu2016deep,song2017deep,jiang2016deep}, many deep hashing methods have been proposed to learn feature representations and hash functions simultaneously \cite{liu2017learning,erin2015deep,zhao2015deep}.
\begin{figure}[t]
\begin{center}
\includegraphics[width=0.5\textwidth,height=2.2in]{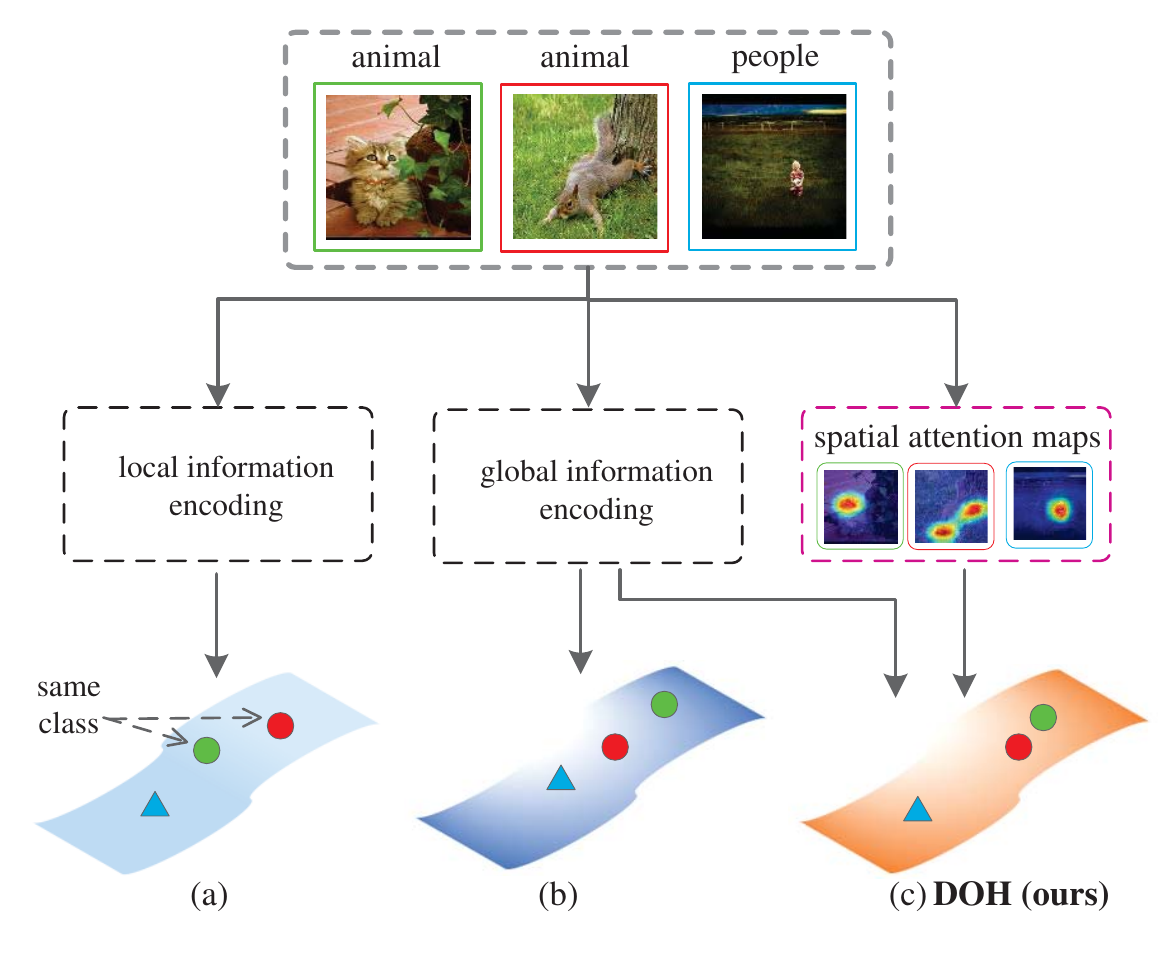}
\end{center}
\caption{Illustration of the idea of the proposed hashing framework. (a) and (b) show the hash codes generated by encoding the local information with FCN network and the global semantic information with CNN network, respectively. (c) shows the proposed hashing method that jointly captures the global semantic structure and the local spatial information with spatial attention maps. The resultant hash codes of the proposed method can achieve accurate matching.}
\label{fig:spatial_example}
\end{figure}

\begin{figure*}[t]
\begin{center}
\includegraphics[width=1\textwidth,height=2.3in]{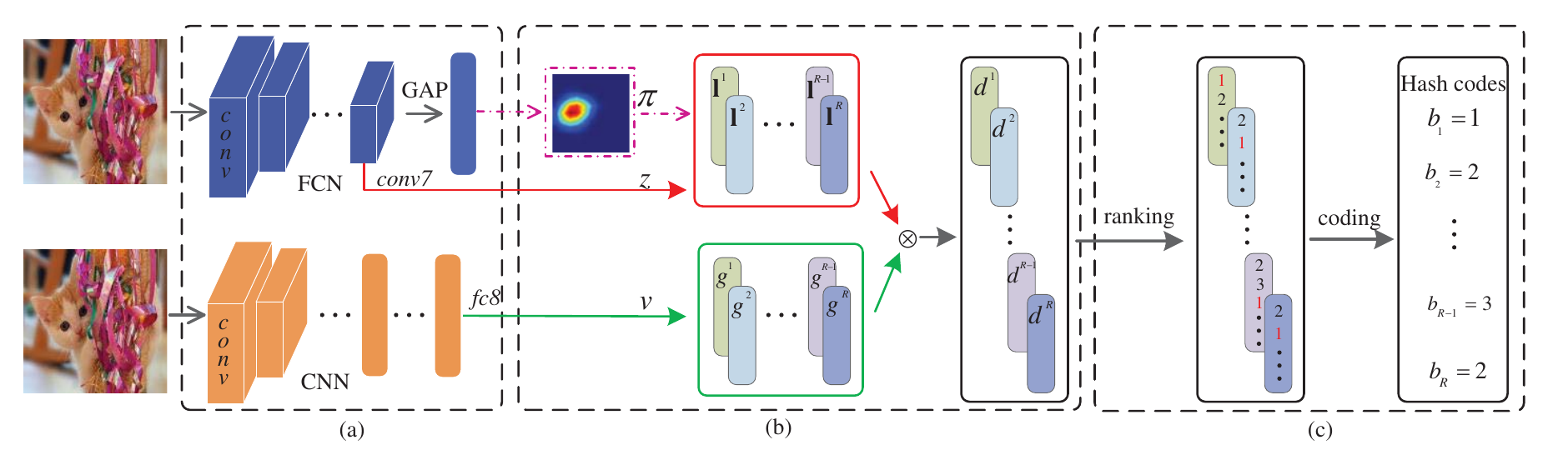}
\end{center}
\caption{Framework of the proposed Deep Ordinal Hashing method, which consists of three components: (a) the feature representation learning which jointly learns the local spatial and global semantic information from FCN and CNN networks, where GAP indicates the global average pooling layer for FCN. (b) a subnetwork to learn the local and global-aware representations $\mathbf{d}^{r}$, where $\odot$ denotes element-wise Hadamard product. In order to capture the local discriminativity, the spatial attention map $\pi$ and the channel-wise representation $\mathbf{z}$ of $conv7$ layer are both leveraged to learn the local-aware representation $\mathbf{l}^{r}$. Additionally, the feature representation $\mathbf{v}$ of $fc8$ layer is used to generate the global-aware representation $\mathbf{g}^{r}$. (c) the ordinal representation learning to approximate the ranking-based hash functions used to generate the hash codes.
}
\label{fig:flowcharthash}
\end{figure*}

As the outputs of the fully-connected layer of CNN have much richer sematic information than hand-crafted features \cite{krizhevsky2012imagenet,oquab2014learning}, most deep hashing methods directly use the outputs of the fully-connected layer to approximate binary hash codes \cite{cao2017deep,erin2015deep,jiang2016deep}. Unfortunately, it may be unsuitable to directly utilize the feature representation from the fully-connected layer due to the spatial information loss \cite{zhou2016learning}, which can lead to the suboptimal hash codes. Besides, recent works on image localization \cite{ren2015faster,oquab2015object} and object detection \cite{li2016weakly,diba2016weakly} have shown the importance of the inherent spatial information from 2-dimensional feature maps of the convolutional layer. Therefore, it is necessary and suitable to explore the local spatial information in the deep hashing framework. By exploring the local spatial and global semantic information simultaneously, the resultant hash codes can preserve ground-truth similarities well, as shown in Figure~\ref{fig:spatial_example}.
On the other hand, as we all know, the binary quantization functions $sign(\cdot)$ and $threshold(\cdot)$ are sensitive to the perturbations in numeric values caused by noise and variations \cite{melucci2007rank}. Meanwhile, ranking-based functions that encode the relative ordering correlation inherit excellent properties of rank correlation measures in terms of scale-invariance, numeric stability, and high nonlinearity \cite{melucci2007rank,broder2000min,yagnik2011power,li2017linear}. Thus, we propose to explore the inherent spatial information to learn the rank correlation space under the deep hashing framework.

In this work, we propose a novel hashing method to explore the global semantic information, the local spatial information and the relative ordering correlation based on the deep learning framework. Here we introduce the convolutional neural network (CNN) \cite{krizhevsky2012imagenet,oquab2014learning} to learn much richer global sematic information than hand-crafted features from images. To explore the local spatial information, an effective spatial attention model is designed to selectively learn well-specified locations closely related to target objects (i.e., aubergine dotted box in Figure~\ref{fig:spatial_example}).
To effectively build the ranking structure, the rank correlation space is learned by exploring the local spatial and global semantic information simultaneously.
By incorporating the above terms into one unified framework, we propose a novel Deep Ordinal Hashing (DOH) method, as illustrated in Figure~\ref{fig:flowcharthash}.
Our network architecture contains three major components: (a) the feature representation learning which learns the local spatial and global semantic information from FCN and CNN respectively; (b) a subnetwork to learn the local and global-aware representation by encoding the local spatial and global semantic information simultaneously; (c) the ordinal representation learning to generate compact hash codes. In summary, we highlight four contributions of this paper as follows:
\begin{itemize}
\item Firstly, we propose a unified framework to learn hash functions by exploring the rank correlation space from both local and global views. To the best of our knowledge, this is the first attempt that learns a group of ranking-based hash functions by jointly exploiting the local spatial and global semantic structure for image retrieval.
\item Secondly, we design a subnetwork to effectively build the rank structure by jointly exploring the local spatial information from FCN and the global semantic information from CNN.
\item Thirdly, we develop an effective spatial attention model to capture the local discriminativity by learning well-specified locations closely related to target objects.
\item Finally, we extensively evaluate the proposed algorithm on three widely-used image retrieval benchmarks. The experimental results show that the proposed algorithm significant outperforms the state-of-the-arts, which demonstrates the superiority and effectiveness of the proposed algorithm.
\end{itemize}

The rest of this paper is organized as follows. In Section 2, we briefly review the related works. The proposed method is introduced in Section 3. We present the optimization algorithm in Section 4. The extensive experiments and discussions of the experimental results are provided in Section 5. Finally, we conclude this work in Section 6.

\section{Related Work}
In this paper, we mainly focus on data-dependent hashing methods that learn hash functions by preserving the data structure. In general, data-dependent hashing methods can be grouped into unsupervised and supervised ones. 
Unsupervised hashing methods usually learn hash functions by exploiting the intrinsic data structure embedded in the original space. For example, Iterative Quantization (ITQ) \cite{gong2013iterative} attempts to generate zero-centered binary codes by maximizing the variance of each binary bit as well as minimizing the quantization error. Representative methods include Spectral Hashing (SH) \cite{weiss2009spectral}, Self-Taught Hashing (STH) \cite{zhang2010self}, Anchor Graph Hashing (AGH) \cite{liu2011hashing}, Neighborhood Discriminant Hashing (NDH) \cite{tang2015neighborhood}, etc.
For supervised methods, the supervised information is incorporated to learn compact binary codes. Existing works have indicated that leveraging the supervised information can produce high quality hash codes. Supervised Hashing with Kernels (KSH) \cite{liu2012supervised} employ a kernel-based functions as the hash functions. Supervised Discrete Hashing (DSH) \cite{shen2015supervised} and Fast Hashing (FastH) \cite{FastHash2015Lin} generate binary hash codes by directly solving the binary programming problem. Other notable methods include Binary Reconstructive Embedding (BRE) \cite{kulis2009learning}, Minimal Loss Hashing (MLH) \cite{norouzi2011minimal}, Two-step Hashing (TSH) \cite{lin2013general}, Asymmetry in Binary Hashing \cite{neyshabur2013power}, Label Preserving Multimedia Hashing (LPMH) \cite{li2017learning}, etc.

Although the aforementioned hashing methods achieve desirable performance, they usually generate suboptimal hash codes due to independently learning feature representations and hash functions. Motivated by the great success of deep learning in computer version tasks, deep hashing frameworks are starting to receive broad attentions recently. Different from conventional hashing methods, deep hashing methods \cite{lai2015simultaneous,zhao2015deep,liong2015deep,zhang2015bit,zhu2016deep,do2016learning,cao2016deep} generate hash codes in such a way that feature representations are optimized during hash functions learning process. The work in \cite{lai2015simultaneous} and \cite{zhao2015deep} is the earliest attempt in jointly learning feature representations and hash functions. Deep Supervised Hashing (DSH) \cite{liu2016deep} tries to leverage similarities and minimize binarization loss simultaneously. Supervised Semantics-preserving Deep Hashing (SSDH) \cite{yang2018supervised} constructs a latent hashing layer to generate hash codes by directly minimizing the classification error on the outputs of the hashing layer.

However, most existing deep hashing methods adopt binary quantization functions, which is known to be sensitive with prevalent noises and variations. In comparison, ranking-based hash functions encode the relative ordering of the projected feature space to generate hash codes, thus benefiting from excellent properties of ordinal measures in scale-invariant, numerically stable, and highly nonlinear \cite{melucci2007rank}. Typical methods contain Min-wise Hash \cite{broder2000min}, Winner-Take-All Hash \cite{yagnik2011power}, Linear Subspace Ranking Hashing \cite{li2017linear} and Supervised Ranking Hashing (SRH) \cite{Li2016Supervised}. By ranking the ordering of feature dimensions represented by the hand-crafted descriptors, these ranking-based hashing methods are inadequate to learn optimal hash functions. Apart from this, the linear transformation of the feature space is insufficient to capture the complex semantic structure of the images, which limits the retrieval performance.
To our best knowledge, this work is the first attempt to learn ranking-based hash functions in an end-to-end manner, where the hash codes are expected to simultaneously capture the local spatial and global semantic information of the images.

\section{Deep Ordinal Hashing}
In this section, we introduce the proposed deep hashing framework in detail, including the representation learning, the spatial attention model and the hash function learning. The entire framework is illustrated in Figure \ref{fig:flowcharthash}.

\subsection{FCN- and CNN-based Representations}
We adopt the widely-used Alexnet as our basic network which includes five convolutional layers from $conv1$ to $conv5$, two fully-connected layers from $fc6$ to $fc7$ and a task-specific fully-connected classification layer (i.e., $fc$-$c$). However, we make some small modification to Alexnet. For CNN network, we add a fully-connected layer (i.e., $fc8$) followed by $fc$-$c$ to learn the feature representations by exploiting the global semantic information. Additionally, for FCN network, we replace $fc6$ and $fc7$ by two convolutional layers (i.e., $conv6$ and $conv7$) followed by the $fc$-$c$ layer, where $conv7$ is followed with a global average pooling layer (i.e., $pool7$). Similar with the CNN network, the $conv7$ layer aims to learn the feature representations by exploiting the local spatial information.

In order to encode the local spatial information and global semantic structure simultaneously, we adopt the feature representations derived from both of the $conv7$ and $fc8$ layers to learn the rank correlation spaces. Formally, we define the channel-wise representation at the spatial location $(x,y)$ of the $conv7$ layer of FCN network as $\mathbf{z}_{xy}$, which can be computed as
\begin{equation}
\mathbf{z}_{xy}=\psi_{\mathcal{F}}(q;\Omega_{\mathcal{F}}),
\end{equation}
where $\mathbf{z}_{xy}\in \mathbb{R}^{M}$ with $M$ being the number of the feature maps, $x \in \{1,\cdots, X\}$, $y \in \{1,\cdots, Y\}$, $X$ and $Y$ are the width and height of feature maps, $q$ presents the input image, $\psi_{\mathcal{F}}$ defines the non-linear projection function for the FCN network and $\Omega_{\mathcal{F}} = \{ \mathbf{W}^{d}_{\mathcal{F}}, \mathbf{b}^{d}_{\mathcal{F}}\}_{d=1}^{D_{\mathcal{F}}}$ defines a set of non-linear projection parameters with $D_{\mathcal{F}}$ being the depth of the FCN network. Additionally, we define the feature representation extracted from the $fc8$ layer of CNN network as $\mathbf{v}$, computed by
\begin{equation}
\mathbf{v} = \psi_{\mathcal{C}}(q;\Omega_{\mathcal{C}}),
\end{equation}
where $\mathbf{v} \in \mathbb{R}^{M}$, $\psi_{\mathcal{C}}$ defines the non-linear projection function for the CNN network and $\Omega_{\mathcal{C}} = \{ \mathbf{W}^{d}_{\mathcal{C}}, \mathbf{b}^{d}_{\mathcal{C}}\}_{d=1}^{D_{\mathcal{C}}}$ defines a set of non-linear projection parameters with $D_{\mathcal{C}}$ being the depth of the CNN network.

\begin{figure*}[t]
\begin{center}
\includegraphics[width=0.9\textwidth]{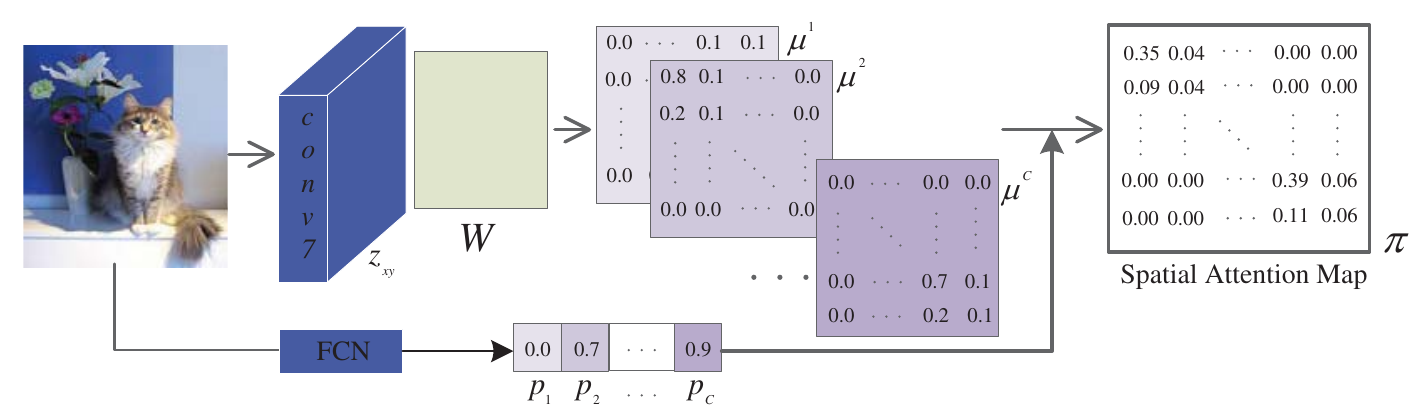}
\end{center}
\caption{An imaginary illustration of the spatial attention model for the FCN network. This model aims to capture the local discriminativity by learning well-specified locations closely related to target objects. Firstly, the object-specific local response map $\mu^{c}$ is obtained by projecting the weight matrix $\mathbf{W}$ of the the classification layer on the channel-wise feature representation $\mathbf{z}_{xy}$ of the $conv7$ layer. After that, the probabilistic outputs $\mathbf{P}$ of the classification layer of the FCN network and object-specific local response maps are jointly leveraged to produce the spatial attention map $\pi$.   }
\label{fig:SpatialRM}
\end{figure*}

\subsection{Spatial Attention Model}
Generally, the target objects of images located in different spatial locations, as shown in Figure~\ref{fig:spatial_example}. For example, when we use the query to search for relevant images in the database, those spatial locations closely related to the target objects are more useful for accurate matching. Instead of considering each spatial location equally, we expect to detect those discriminative spatial locations that are highly related to the objects and give larger response values to them. Consequently, we introduce a spatial attention model, which aims to generate object-specific spatial attention map. Figure~\ref{fig:SpatialRM} illustrates the generation of the spatial attention map. We also visualize the spatial attention maps of some examples in the purple dotted box in Figure~\ref{fig:spatial_example}.

As mentioned above, we perform the global average pooling on the $conv7$ layer and feed the outputs into the fully-connected classification layer (i.e.,$fc$-$c$). The outputs of the global average pooling can be regarded as the spatial average of feature maps of the $conv7$ layer. Those spatial average values are used to generate the probabilistic outputs of the $fc$-$c$ layer. In this section, we introduce an intuitive way to produce the spatial attention map by projecting the weight matrix of the $fc$-$c$ layer on the feature maps of the $conv7$ layer. Consider the weight matrix $\mathbf{W} \in \mathbb{R}^{M \times C}$ performs a mapping of the spatial average values to the semantic class labels, where $C$ defines the total number of categories. In particular, we define the object-specific local response at the spatial location $(x,y)$ as $\mu_{xy}^{c}$, which can be obtained by
\begin{equation}
\mu_{xy}^{c} = \max (\mathbf{w}_{c}^{T}\mathbf{z}_{xy},0),\ \text{for} \ c = 1,\cdots,C,
\end{equation}
where $\mathbf{w}_{c}$ is the $c$-th column of $\mathbf{W}$. Obviously, $\mu_{xy}^{c}$ indicates the importance at the spatial location $(x,y)$ that an image is classified to the $c$-th class. As each spatial location $(x,y)$ corresponds to different local patch in the original image, the local response $\mu_{xy}^{c}$ indicates the relative similarity of the local image patch to the $c$-th class.

Essentially, we can obtain the local discriminative information at different spatial locations for a particular class. The bigger value of $\mu_{xy}^{c}$ indicates that the image patch at the spatial location $(x,y)$ is more relative to the $c$-th class. Inversely, the smaller value of $\mu_{xy}^{c}$ indicates less relative to the $c$-th class. By integrating $\mu_{xy}^{c}$ together, we can define the spatial attention map $\pi$ to identifying all the object-specific image patches. Specifically, the local response at $(x,y)$, denoted as $\pi_{xy}$, is defined as follow
\begin{equation}
\pi_{xy}= \frac{{\sum\limits_{c=1}^{C} p_{c}\mu_{xy}^{c}}}{{ \sum\limits_{c=1}^{C}p_{c}}},
\label{pi}
\end{equation}
where $p_{c}$ is the $c$-th probabilistic outputs of the classification (i.e.,$fc$-$c$) layer.

\subsection{Local and Global-aware Representations}
In order to jointly capture the local spatial and global semantic information to learn ranking-based hash functions, we require the ordinal representations which are used to approximate the hash functions to be local and global aware. As shown in Figure~\ref{fig:flowcharthash}, DOH jointly encodes the feature representations of the $conv7$ layer and spatial attention map to generate the local-aware representations which explicitly exploits the local spatial structure of images. In addition, DOH learns global-aware representations by encoding the global semantic information from the feature representations of the $fc8$ layer. Let $\mathbf{l},\mathbf{g}\in \mathbb{R}^{K}$ be the local-aware and global-aware representations respectively. To jointly preserve both of local and global information, we further integrate them to define the confident score $d_{k}$ as follow
\begin{equation}
d_k = l_{k}g_{k},\ \text{for} \ k = 1,\cdots,K,
\label{dk}
\end{equation}
where $l_{k}$ and $g_{k}$ are the $k$-th dimension of $\mathbf{l}$ and $\mathbf{g}$ respectively. By connecting $d_k$ together, we can define $\mathbf{d}=[d_{1},\cdots,d_{K}]\in \mathbb{R}^{K}$ as the local and global-aware representations.

\begin{figure*}[t]
\begin{center}
\includegraphics[width=1\textwidth]{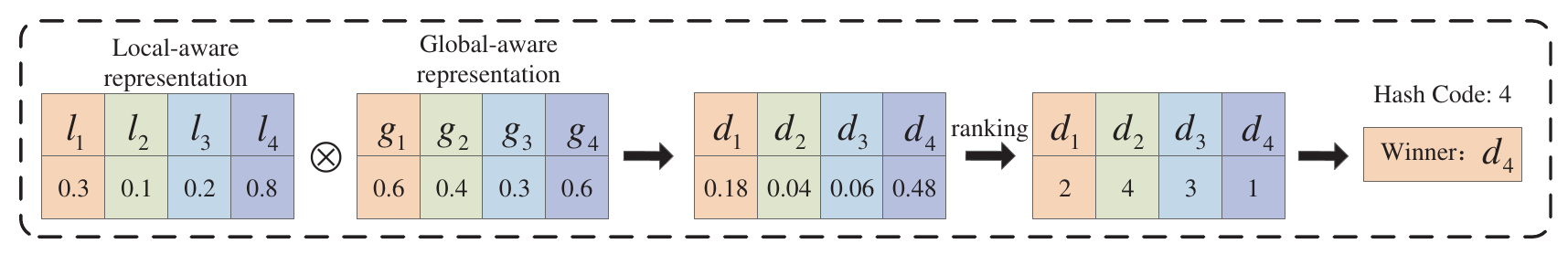}
\end{center}
\caption{An example with 4-dimensional feature space to produce $4$-ary hash code.}
\label{fig:DTARanking}
\end{figure*}

\subsubsection{Local-aware Representations}
 As illustrated in Figure~\ref{fig:SpatialRM}, the spatial attention map demonstrates the local discriminativity in identifying partial image regions that are highly related to target objects. When searching relevant images in the database, those discriminative spatial locations are crucial important for accurate matching. We argue for learning above local spatial structure for ordinal representations by jointly encoding the discriminative spatial attention map and convolutional feature maps of the $fc7$ layer. Firstly, we define the transformation matrix and the bias vector as $\mathbf{W}_{s}= [\mathbf{w}_{s1},\cdots,\mathbf{w}_{sK}] \in \mathbb{R}^{M \times K}$ and $\mathbf{b}_{s} \in \mathbb{R}^{K}$ respectively, which are used to project $\mathbf{z}_{xy}$ into local-ware representations $\mathbf{l}$. For each latent patten $\mathbf{w}_{sk}$\footnote {As $\mathbf{w}_{sk}$ and $\mathbf{w}_{gk}$ are unlabeled, we consider them as latent.\label{fn:repeat}}, we define $\omega_{xy}^{k}$ as the spatial confident score over the occurrence of $\mathbf{w}_{sk}$ at the spatial location $(x,y)$, which is given by,
\begin{equation}
\omega_{xy}^{k} = \mathbf{w}_{sk}^{T}\mathbf{z}_{xy}+b_{sk},
\label{omegaxy}
\end{equation}
where $\mathbf{w}_{sk}$ is the $k$-th column of $\mathbf{W}_{s}$, $b_{sk}$ is the $k$-th element of $\mathbf{b}_{s}$.

To model the local spatial structure, we employ the softmax function to calculate the probability that the latent patten $\mathbf{w}_{sk}$ appears at the location $(x,y)$ as follow
\begin{equation}
\xi_{x,y}^{k}= \frac{{{\exp(\omega_{xy}^{k})}}}{{ {{\sum\nolimits_{x',y'=1}^{X,Y} {\exp (\omega_{x'y'}^{k})}}} }}.
\end{equation}
Furthermore, we leverage spatial attention map to estimate the local awareness $l_{k}$ carried by the latent pattern $\mathbf{w}_{sk}$ for identifying object-specific local regions. In particulary, we define $l_{k}$ as follow
\begin{equation}
l_{k}= \sum\limits_{x,y} {\pi_{xy}\xi_{xy}^{k}}.
\label{lk}
\end{equation}
By concatenating $l_{k}$ together, we can obtain the local-aware representations as $\mathbf{l}=[l_{1},\cdots,l_{K}]$, $\phi(a) = (1+\exp(-a))^{-1}$ defines the sigmoid function.

\subsubsection{Global-aware Representations} In addition to exploit the local spatial structure, the ordinal representation is still expected to preserve the global semantic information from the visual descriptors of the $fc8$ layer. Let $\mathbf{W}_{g}= [\mathbf{w}_{g1},\cdots,\mathbf{w}_{gK}] \in \mathbb{R}^{M \times K}$ and $\mathbf{b}_{g} \in \mathbb{R}^{K}$ be the transformation matrix and the bias vector, which project the global feature representation of CNN network into global-aware representations.
Similarly, we define the global awareness $g_{k}$ as follow,
\begin{equation}
g_{k} = \mathbf{w}_{gk}^{T}\mathbf{v}+b_{gk},
\label{gk}
\end{equation}
where $\mathbf{w}_{gk}$ is the $k$-th column of $\mathbf{W}_{g}$ and $b_{gk}$ is the $k$-th element of $\mathbf{b}_{g}$. Intuitively, the global awareness $g_{k}$ carried by the latent pattern $\mathbf{w}_{gk}$\footref{fn:repeat} measures the relative similarity to the image from the global perspective. By concatenating $g_{k}$ together, we can obtain the global-aware representations as $\mathbf{g}=[g_{1},\cdots,g_{K}]$.

\subsection{Ranking-based Hash Functions}
The proposed DOH method targets to learn ranking-based hash functions by encoding the local spatial and global semantic information from deep networks. In this work, we develop an intuitive ranking-based hash functions by encoding the comparative ordering of the local and global-aware representations. Figure~\ref{fig:DTARanking} shows an example with 4-dimensional feature space which is encoded to generate one hash bit. DOH firstly generates the representation $\mathbf{d}$ from the local-aware representation $\mathbf{l}$ and the global-aware representation $\mathbf{g}$.
After that, a permutation on $\mathbf{d}$ is performed in descending order. The dimension (i.e., $d_{4}$ in Figure~\ref{fig:DTARanking}) which takes the maximum value will win the comparison and its index will be used as the hash code (i.e., 4). In the following, we give the mathematical formulations of ranking-based hashing functions in detail.

Suppose $\mathbf{d}^{r}=[d_{1}^{r},\cdots,d_{K}^{r}]$ is the local and global-aware representation which is used to learn the $r$-th hash code.
We define $f^{r}(\cdot)$ as the ranking-based hash function for the $r$-th hash code, which can be calculated as
\begin{equation}
\begin{array}{l}
f^{r}(\mathbf{z},\mathbf{v};\mathbf{W}_{s}^{r},\mathbf{b}_{s}^{r},\mathbf{W}_{g}^{r},\mathbf{b}_{g}^{r})=arg\mathop {max}\limits_{\theta}\theta^{T} \mathbf{d}^{r}  \\
 s.t. \ \ \theta \in \{0,1\}^{K}, \theta^{T}\mathbf{1}=1,\\
 \end{array}
 \label{h}
\end{equation}
where $\mathbf{W}_{s}^{r}=[\mathbf{w}_{s1}^{r},\cdots,\mathbf{w}_{sK}^{r}]$ and $\mathbf{b}_{s}^{r}=[b_{s1}^{r},\cdots, b_{sK}^{r}]$ define the transformation matrix and the bias vector of FCN network for generating the $r$-th hash code, $\mathbf{W}_{g}^{r}=[\mathbf{w}_{g1}^{r},\cdots,\mathbf{w}_{gK}^{r}]$ and $\mathbf{b}_{g}^{r}=[b_{g1}^{r},\cdots,b_{gK}^{r}]$ define the transformation matrix and the bias vector of CNN network for generating the $r$-th hash code, $\mathbf{1}$ is a vector with each element being 1, and $\mathbf{d}^{r}$ can be calculated in terms of Eq.~(\ref{dk}). Meanwhile, the constraints in Eq.~(\ref{h}) act as an 1-of-$K$ indicator of the rank ordering of the input representation $\mathbf{d}^{r}$. \textbf{Algorithm 1} summarizes the entire DOH hashing procedure to produce a code sequence $\mathbf{b}$.

\begin{algorithm}[t]
\caption{Deep Ordinal Hashing}
\label{alg1}
\begin{algorithmic}
\REQUIRE The input image $q$, code length $R$, network parameter sets $\Omega_{\mathcal{F}}$, $\Phi_{\mathcal{F}}=\{\mathbf{W}_{s}^{r},\mathbf{b}_{s}^{r}\}_{r=1}^{R}$, $\Omega_{\mathcal{C}}$, $\Phi_{\mathcal{C}}=\{\mathbf{W}_{g}^{r},\mathbf{b}_{g}^{r}\}_{r=1}^{R}$.
\ENSURE Hash code $\mathbf{b}=[b_{1},\cdots,b_{r},\cdots,b_{R}]$.
\STATE Compute $\mathbf{z}_{xy}=\psi_{\mathcal{F}}(q;\Omega_{\mathcal{F}})$ and $\mathbf{v} = \psi_{\mathcal{C}}(q;\Omega_{\mathcal{C}})$ by forward propagation.
\STATE Calculate the spatial attention map $\pi$ according to Eq.~(\ref{pi}).
\FOR{$r=1,\cdots,R$}
\STATE Calculate the local-ware representation $\mathbf{l}$ and the global-ware representation $\mathbf{g}$ according to Eq.~(\ref{omegaxy})-(\ref{lk}) and Eq.~(\ref{gk}), respectively.
\STATE Calculate the local and global-aware representation $\mathbf{d}=[d_{1},\cdots,d_{K}]$ according to Eq.~(\ref{dk}).
\STATE $b_{r}=\hat{k} \leftarrow \arg \mathop {\max }\limits_{1 \le k \le K} d_{k}$.
\ENDFOR
\end{algorithmic}
\end{algorithm}

It is worth to note that the $arg\mathop {max}$ term in Eq.~(\ref{h}) is non-convex and highly discontinuous, thus making it hard to optimize. To make it tractable, the softmax function is employed to approximate Eq.~(\ref{h}). Therefore, we reformulated Eq.~(\ref{h}) and define the ordinal representation $\mathbf{h}^{r}$ as follow
\begin{equation}
\mathbf{h}^{r}(\mathbf{z},\mathbf{v};\mathbf{W}_{s}^{r},\mathbf{b}_{s}^{r},\mathbf{W}_{g}^{r},\mathbf{b}_{g}^{r})=\text{softmax}( \mathbf{d}^{r}),
\end{equation}
where $\mathbf{h}^{r}=[h_{1}^{r},\cdots,h_{K}^{r}]$ is the probabilistic approximation of the hash function $f^{r}$. In fact, the entry $h_{k}^{r}$ of $\mathbf{h}^{r}$ represents the probability of the $k$-th dimension taking the maximum value of $\mathbf{d}^{r}$.
Specifically, the probability $h_{k}^{r}$ can be calculated as
\begin{equation}
h_{k}^{r}= \frac{{\exp (d_{k}^{r})}}{{\sum\limits_{k'=1}^{K} {\exp (d_{k'}^{r})}}}, \text{for} \ \  k=1,\cdots,K,
\label{hk}
\end{equation}
where $h_{k}^{r}$ can be interpreted as the probability that both of the latent patterns $\mathbf{w}_{s}^{r}$ and $\mathbf{w}_{g}^{r}$ contain the most discriminative information.


Suppose an image pair $(q_{i},q_{j})$ with their corresponding similarity label $s_{ij}\in \{0,1\}$ is given, we try to learn a set of hash functions which makes the hash codes of the similar pairs close but dissimilar pairs apart. Let $\mathbf{b}(i)$ and $\mathbf{b}(j)$ be the code sequence for the image $q_{i}$ and $q_{j}$ respectively.
Formally, we define the probability that the $k$-th dimension of the feature space is selected as the winner for the $r$-th hash bit of both $\mathbf{b}(i)$ and $\mathbf{b}(j)$ as $\varphi_{r}^{ij}$, which can be calculated by
\begin{equation}
\begin{array}{l}
 \varphi_{r}^{ij} \equiv P(b_{r}(i)=b_{r}(j) | \mathbf{W}_{s}^{r},\mathbf{b}_{s}^{r},\mathbf{W}_{g}^{r},\mathbf{b}_{g}^{r}) \\
  \quad \ \  = \sum\limits_{k=1}^{K} {P(b_{r}(i)=k | \mathbf{W}_{s}^{r},\mathbf{b}_{s}^{r},\mathbf{W}_{g}^{r},\mathbf{b}_{g}^{r} )} \\
  \quad \quad \ \ \ \ \ \ \ {P(b_{r}(j)=k | \mathbf{W}_{s}^{r},\mathbf{b}_{s}^{r},\mathbf{W}_{g}^{r},\mathbf{b}_{g}^{r} )}  \\
  \quad \ \  = \sum\limits_{k=1}^{K} {h_{k}^{r}(i) h_{k}^{r}(j)} = (\mathbf{h}^{r}(i))^{T}\mathbf{h}^{r}(j), \\
 \end{array}
 \label{Prij}
\end{equation}
where $b_{r}(i)$ and $b_{r}(j)$ are the $r$-th bit of $\mathbf{b}_{r}(i)$ and $\mathbf{b}_{r}(j)$ respectively, and $\mathbf{h}^{r}(i)$ and $\mathbf{h}^{r}(j)$ are the softmax vectors corresponding to the $r$-th hash code of the image $q_{i}$ and $q_{j}$ respectively.

Meanwhile, we also expect to learn the entire code sequence jointly so as to exploit the complementary information of each individual hash bit. In particular, we define $\varepsilon_{ij} $ as the probability that two code sequence $\mathbf{b}(i)$ and $\mathbf{b}(j)$ take the same values. Here, we can obtain $\varepsilon_{ij} $ by
\begin{equation}
\begin{array}{l}
 \varepsilon_{ij} \equiv P (\mathbf{b}(i)=\mathbf{b}(j) |  \Phi_{\mathcal{F}}, \Phi_{\mathcal{C}} )  \\
  \quad \  = \sum\limits_{r=1}^{R}P(b_{r}(i)=b_{r}(j) |\mathbf{W}_{s}^{r},\mathbf{b}_{s}^{r},\mathbf{W}_{g}^{r},\mathbf{b}_{g}^{r} )/\sum\limits_{r=1}^{R} {\sum\limits_{k=1}^{K} {h_{k}^{r}} } \\
  \quad \   =\frac{1}{R} \sum\limits_{r=1}^{R} {\varphi_{r}^{ij}}, \\
 \end{array}
\end{equation}
where $\Phi_{\mathcal{F}} = \{ \mathbf{W}_{s}^{r},\mathbf{b}_{s}^{r} \}_{r=1}^{R} $ and $\Phi_{\mathcal{C}} = \{\mathbf{W}_{g}^{r},\mathbf{b}_{g}^{r} \}_{r=1}^{R} $ denotes all $R$ transformation matrixes and bias vectors for the FCN and CNN network respectively.

Actually, the larger value of $\varepsilon_{ij} $ indicates that the code sequence $\mathbf{b}(i)$ and $\mathbf{b}(j)$ are more similar with each other. But conversely, the smaller value of $\varepsilon_{ij} $ indicates less similar of two code sequence. That is, if $s_{ij}=1$, $\varepsilon_{ij} $ should be pushed towards 1, otherwise $\varepsilon_{ij} $ should be pushed to 0. Intuitively, our goal is to maximize $\varepsilon_{ij} $ when $s_{ij}=1$ and minimize $\varepsilon_{ij} $ when $s_{ij}=0$. This inspires us to employ the Euclidian loss function to define the following objective function
\begin{equation}
\ell_{ij}(q_{i},q_{j},s_{ij})=\frac{1}{2}(\varepsilon_{ij}-s_{ij})^{2}.
\label{lij}
\end{equation}
Let $\Gamma = \{(q_{i},q_{j}),s_{ij}\}_{i,j=1}^{N}$ be the training set. We define the overall loss function over the training set $\Gamma$ as follow
\begin{equation}
\mathcal{L} (\Gamma;\Omega_{\mathcal{F}},\Phi_{\mathcal{F}},\Omega_{\mathcal{C}},\Phi_{\mathcal{C}}) = \frac{1}{N}\sum\limits_{s_{ij} \in \mathbf{S}} {\ell_{ij}(q_{i},q_{j},s_{ij})}.
\label{L}
\end{equation}
Finally, the proposed framework can be formulated as
\begin{equation}
\mathop {\min }\limits_{\Omega_{\mathcal{F}},\Phi_{\mathcal{F}},\Omega_{\mathcal{C}},\Phi_{\mathcal{C}}} \mathcal{L}(\Gamma;\Omega_{\mathcal{F}},\Phi_{\mathcal{F}},\Omega_{\mathcal{C}},\Phi_{\mathcal{C}}).
\end{equation}

\begin{algorithm}[t]
\caption{The learning algorithm for DOH}
\label{alg2}
\begin{algorithmic}
\REQUIRE Training set $\Gamma$.
\ENSURE Network parameter sets $\Omega_{\mathcal{F}}$, $\Phi_{\mathcal{F}}$, $\Omega_{\mathcal{C}}$, $\Phi_{\mathcal{C}}$. 
\STATE \textbf{Initialization} \\
Initialize Network parameter sets $\Omega_{\mathcal{F}}$, $\Phi_{\mathcal{F}}$, $\Omega_{\mathcal{C}}$, $\Phi_{\mathcal{C}}$, iteration number $N_{iter}$ and mini-batch size $N_{batch}$.
\STATE \textbf{repeat:}
\FOR{$iter = 1, \cdots,N_{iter}$}
\STATE Randomly select $N_{batch}$ image pairs from $\Gamma$.
\FOR{$r=1,\cdots,R$}
\STATE For each selected image pair, compute $\mathbf{h}^{r}$ by forward propagation according to Eq.~(\ref{hk}).
\STATE Calculate the derivatives $\frac{\partial \mathcal{L} }{\partial \mathbf{W}_{s}^{r} }$, $\frac{\partial \mathcal{L} }{\partial \mathbf{b}_{s}^{r} }$, $\frac{\partial \mathcal{L} }{\partial \mathbf{W}_{g}^{r} }$ and $\frac{\partial \mathcal{L} }{\partial \mathbf{b}_{g}^{r} }$ according to Eq.~(\ref{gradLUr}),~(\ref{gradLbr}),~(\ref{gradlUr}),~(\ref{gradVareUr}),~(\ref{gradVarphiUr}),~(\ref{gradHUr}),~(\ref{gradd}),
~(\ref{graddg}).
\STATE Update the parameters $\mathbf{W}_{s}^{r}$, $\mathbf{b}_{s}^{r}$, $\mathbf{W}_{g}^{r}$, $\mathbf{b}_{g}^{r}$ using the BP algorithm.
\ENDFOR
\STATE Update parameter sets $\Omega_{\mathcal{F}} $ and $\Omega_{\mathcal{C}}$ for the FCN and CNN network, respectively.
\ENDFOR
\STATE \textbf{until} a fixed number of iterations.
\end{algorithmic}
\end{algorithm}

\section{Optimization}
The proposed DOH algorithm involves four sets of variables, such as the parameter sets $\Omega_{\mathcal{F}} $ and $\Phi_{\mathcal{F}}$ for the FCN network and the parameter sets $\Omega_{\mathcal{C}}$ and $\Phi_{\mathcal{C}}$ for the CNN network. As it is non-convex to simultaneously learn $\Omega_{*}$ and $\Phi_{*}$, to make it tractable, we adopt an alternating optimization approach to update one variable by fixing the rest variables, where $*$ is a placeholder for $\mathcal{F}$ and $\mathcal{C}$. The proposed optimization algorithm is summarized in \textbf{Algorithm 2}. In the following, we discuss the derivatives of $\mathcal{L}$ in detail.

\textbf{Update $\Phi_{\mathcal{F}}$.} We first solve $\Phi_{\mathcal{F}}$ with the parameter sets $\Phi_{\mathcal{C}}$, $\Omega_{\mathcal{C}}$ and $\Omega_{\mathcal{F}}$ fixed. As $\mathbf{W}_{s}^{r}$ and $\mathbf{b}_{s}^{r}$ are independent on $\{ \mathbf{W}_{s}^{r'},\mathbf{b}_{s}^{r'}\}_{r'\neq r}$, we can optimize $\Phi_{\mathcal{F}}$ by decomposing it to $R$ independent subproblems, where each subproblem can be formulated as the following optimization problem:
\begin{equation}
\mathop {\min }\limits_{\mathbf{W}_{s}^{r},\mathbf{b}_{s}^{r}} \mathcal{L}(\Gamma;\Omega_{\mathcal{F}},\Phi_{\mathcal{F}},\Omega_{\mathcal{C}},\Phi_{\mathcal{C}}).
\label{JUr}
\end{equation}
The optimization problem in Eq.~(\ref{JUr}) can be efficient solved by using the stochastic gradient descent (SGD) with the back-propagation (BP) algorithm. The gradient of the objective function in Eq.~(\ref{JUr}) $\text{w.r.t.}$ $\mathbf{W}_{s}^{r}$ and $\mathbf{b}_{s}^{r}$ can be computed as follow
\begin{equation}
\frac{\partial \mathcal{L} }{\partial \mathbf{W}_{s}^{r} }=\frac{1}{N}\sum\limits_{s_{ij} \in \mathbf{S}} { \frac{\partial \ell_{ij} }{\partial \mathbf{W}_{s}^{r} }},
\label{gradLUr}
\end{equation}
\begin{equation}
\frac{\partial \mathcal{L} }{\partial \mathbf{b}_{s}^{r} }=\frac{1}{N}\sum\limits_{s_{ij} \in \mathbf{S}} { \frac{\partial \ell_{ij} }{\partial \mathbf{b}_{s}^{r} }}.
\label{gradLbr}
\end{equation}
Specifically, the gradient $\frac{\partial \ell_{ij} }{\partial  \mathbf{W}_{s}^{r} }$ in Eq.~(\ref{gradLUr})
can be calculated by using the chain rule of the derivatives on $\ell_{ij}$ in Eq.~(\ref{lij}), which can be calculated as follow
\begin{equation}
\frac{\partial \ell_{ij} }{\partial \mathbf{W}_{s}^{r} }=(\varepsilon_{ij}-s_{ij})\frac{\partial \varepsilon_{ij}, }{\partial \mathbf{W}_{s}^{r} }
\label{gradlUr}
\end{equation}
\begin{equation}
\frac{\partial \varepsilon_{ij} }{\partial \mathbf{W}_{s}^{r} }= \frac{1}{R} \frac{\partial \varphi_{r}^{ij} }{\partial \mathbf{W}_{s}^{r} },
\label{gradVareUr}
\end{equation}
\begin{equation}
\begin{array}{l}
 \frac{\partial \varphi_{r}^{ij} }{\partial \mathbf{W}_{s}^{r} } = (\mathbf{h}^{r}(i) \odot \mathbf{h}^{r}(j)- (\mathbf{h}^{r}(i))^{T} \mathbf{h}^{r}(j) \mathbf{h}^{r}(i)) \frac{\partial \mathbf{d}^{r}(i) }{\partial \mathbf{W}_{s}^{r} }  \\
  \quad \quad \ \ + (\mathbf{h}^{r}(j)\odot  \mathbf{h}^{r}(i)- (\mathbf{h}^{r}(j))^{T} \mathbf{h}^{r}(i) \mathbf{h}^{r}(j)) \frac{\partial \mathbf{d}^{r}(j) }{\partial \mathbf{W}_{s}^{r} }, \\
 \end{array}
\label{gradVarphiUr}
\end{equation}
\begin{equation}
\frac{\partial \mathbf{d}^{r}}{\partial \mathbf{W}_{s}^{r} } = [ \frac{\partial d_{1}^{r}}{\partial \mathbf{w}_{s1}^{r} },\cdots, \frac{\partial d_{k}^{r}}{\partial \mathbf{w}_{sk}^{r} },\cdots,  \frac{\partial d_{K}^{r}}{\partial \mathbf{w}_{sK}^{r} }],
\label{gradHUr}
\end{equation}
\begin{equation}
\frac{\partial d_{k}^{r}}{\partial \mathbf{w}_{sk}^{r} }=l_{k}^{r}g_{k}^{r}[\mathbf{z}_{xy}^{T} - \sum\limits_{x'y'} {\xi_{x'y'}^{kr}} \mathbf{z}_{x'y'}^{T} ],
\label{gradd}
\end{equation}
where $\odot$ stands for the element-wise Hadamard product. As the gradient $ \frac{\partial \ell_{ij} }{\partial \mathbf{b}_{s}^{r} }$ is similar with $\frac{\partial \ell_{ij} }{\partial \mathbf{W}_{s}^{r} }$, we do not elaborate its solution here.


\textbf{Update $\Phi_{\mathcal{C}}$.} Actually, the solution of $\Phi_{\mathcal{C}}$ is similar with that of $\Phi_{\mathcal{F}}$ except for the terms in Eq.~(\ref{gradd}). Therefore, we only calculate the derivatives of $d_{k}^{r}$ as follow
\begin{equation}
\frac{\partial d_{k}^{r}}{\partial \mathbf{w}_{gk}^{r} }=l_{k}^{r}\mathbf{v}^{T}
\label{graddg}
\end{equation}

\textbf{Update $\Omega_{*}$.} The parameters $\mathbf{W}^{d}_{*}$ and $\mathbf{\mathbf{b}^{d}_{*}}$ can be  automatically updated by applying SGD with BP algorithm in Caffe \cite{jia2014caffe}.

\section{Experiments}
In this section, we conduct extensive experiments to verify the effectiveness of DOH on three widely-used image retrieval datasets including MIRFlickr25k \cite{huiskes2008mir}, CIFAR-10 \cite{babenko2014additive} and NUS-WIDE \cite{chua2009nus}.

\subsection{Datasets}
\textbf{MIRFlickr25k}\footnote {http://press.liacs.nl/mirflickr/ \label{fn:flickr}} contains 25,000 images collected from Flickr website. In this dataset, each image is annotated with one or more of the 24 ground truth semantic labels. We randomly select 908 images from this dataset as queries and the rest are used to form the database, from which we randomly sample 5000 images to form the training set.

\textbf{CIFAR-10}\footnote {http://www.cs.toronto.edu/kriz/cifar.html \label{fn:cifar}} includes 60,000 real world tiny images belonging to 10 classes, where each category has 6,000 images. We randomly select 100 images per class as the queries, 500 images per class as the training set and the rest forms the
database.

\textbf{NUS-WIDE}\footnote {http://lms.comp.nus.edu.sg/research/NUS-WIDE.htm \label{fn:nus}} consists of 269,648 images crawled from Flickr websit, where each image is associated with one or more of the 81 ground truth sematic concepts (labels). For this dataset, we manually select 195,834 images belonging to the 21 largest concepts. We randomly select 100 images per class as the queries and the rest images form the databese, from which we select 500 images per class as the training set.

\subsection{Baselines and Evaluation Metrics}
We have compared the proposed DOH method with several state-of-the-art image hashing methods, including five non-deep hashing methods (i.e., LSH \cite{andoni2006near}, ITQ \cite{gong2013iterative}, KSH \cite{liu2012supervised}, SDH \cite{shen2015supervised} and FastH \cite{FastHash2015Lin}) and three deep hashing methods (i.e., DNNH \cite{lai2015simultaneous}, DSH \cite{liu2016deep} and SSDH \cite{yang2018supervised}). We have briefly introduced those hashing methods in Section II. For non-deep methods we extract the $4096$-dimensional feature of $fc7$ layer using the Alexnet network. For all deep methods, we adopt the same network (i.e., Alexnet) for fair comparison. Except for DNNH, the source codes of the baselines are kindly provided by their authors. Specifically, we implement DNNH method with the open-source Caffe \cite{jia2014caffe} framework. The parameters for all the compared methods are selected by their default ones. In addition, we evaluate the retrieval performance using three widely used metrics: mean Average Precision (mAP), top-N Precision (P@N) and Precision-Recall curves (PR). 

\begin{table}[t]
\begin{center}
\caption{Configurations of the two-stream network}
\label{tab:network}
     \begin{tabular}{|l|l|p{5.5cm}|}
        \hline
        Network&Layer & Configuration \\
        \hline
        \hline
        \multirow{14}{*}{FCN}
        & \multirow{2}{*}{$conv1$} & Filter: $96 \times 11 \times 11$; Pad: 0; Stride:4;\\
        & &ReLU; Pool; LRN; \\
        &\multirow{2}{*}{$conv2$} & Filter: $256 \times 5 \times 5$; Pad: 2; Stride:1; \\
        & & ReLU; Pool; LRN;   \\
        &\multirow{2}{*}{$conv3$} & Filter: $ 384\times 3 \times 3$; Pad: 1; \\
        & & Stride:1; ReLU;    \\
        &\multirow{2}{*}{$conv4$} & Filter: $384 \times 3 \times 3$; Pad: 1; \\
        & & Stride:1; ReLU;  \\
        &\multirow{2}{*}{$conv5$} & Filter: $384 \times 3 \times 3$; Pad: 1; \\
        & & Stride:1; ReLU; Pool;  \\
        &$conv6$ & Filter: $512 \times 3 \times 3$; Pad: 1; ReLU;  \\
        &$conv7$ & Filter: $512 \times 3 \times 3$; Pad: 1; ReLU; Pool; Dropout;\\
        &$fc$-$c$ & Number: $C$;    \\
        &$fc$-$h$ & Number: $ K \times R$;     \\
        \hline
         \multirow{15}{*}{CNN}
        &\multirow{2}{*}{$conv1$} & Filter: $96 \times 11 \times 11$; Pad: 0; Stride:4;\\
        & & ReLU; Pool; LRN; \\
        &\multirow{2}{*}{$conv2$} & Filter: $256 \times 5 \times 5$; Pad: 2; Stride:1; \\
        & & ReLU; Pool; LRN;   \\
        &\multirow{2}{*}{$conv3$} & Filter: $ 384\times 3 \times 3$; Pad: 1; \\
        & &Stride:1; ReLU;    \\
        &\multirow{2}{*}{$conv4$} & Filter: $384 \times 3 \times 3$; Pad: 1; \\
        & &Stride:1; ReLU;  \\
        &\multirow{2}{*}{$conv5$} & Filter: $384 \times 3 \times 3$; Pad: 1; \\
        & &Stride:1; ReLU; Pool;  \\
        &$fc6$ & Number: $4096$; ReLU; Dropout;  \\
        &$fc7$ & Number: $4096$; ReLU; Dropout;  \\
        &$fc8$ & Number: $512$; ReLU; Dropout;    \\
        &$fc$-$c$ & Number: $C$;    \\
        &$fc$-$h$ & Number: $ K \times R$;    \\
     \hline
     \specialrule{0em}{1pt}{1pt}
     \end{tabular}
\end{center}
\end{table}

\begin{table*}[t]
\begin{center}
\caption{mAP results of all methods with respect to different code length on three datasets. }
\label{tab:map}
\begin{tabular}{ |c|cccc|cccc|cccc| }
\hline
  \multirow{2}{*}{Method}&
 \multicolumn{4}{c|}{MIRFLICKR25K} & \multicolumn{4}{c|}{CIFAR-10}& \multicolumn{4}{c|}{NUS-WIDE}\\
 \cline{2-13}
  & 8 bits & 16 bits & 24 bits&32 bits& 8 bits & 16 bits & 24 bits&32 bits& 8 bits & 16 bits & 24 bits&32 bits \\
 \hline
 \hline
 LSH  &0.5721 & 0.5845& 0.5826 & 0.5863& 0.1087&0.1137 &0.1172 &0.1331 & 0.3845& 0.4047& 0.4090& 0.4092\\
 ITQ  &0.6448 &0.6472 &0.6515  &0.6517 & 0.1767& 0.1839&0.1858 &0.1902 &0.4725 &0.4835 &0.4880 &0.4943 \\
 KSH &0.6843 &0.6968 &0.7001  &0.7035 &0.3096 &0.3599 & 0.3766&0.3892 & 0.4985& 0.5084& 0.5160&0.5236\\
 SDH  &0.7268 &0.7292 &0.7347 & 0.7416&0.3371 & 0.4779& 0.5142&0.5200 & 0.5221&0.5394 &0.5415&0.5488\\
 FastH  &0.7453 &0.7748 &0.7909 &0.7970& 0.4253& 0.4909&0.5151 &0.5395 &0.5339 & 0.5570&0.5732&0.5822\\
 \hline
  DNNH  & 0.7498 &0.7622 & 0.7745 &0.7670 &0.5803 &0.5959 &0.6329 &0.6358 & \underline{0.6472}&0.6598 &0.6747 &0.6784 \\
  DSH   & 0.7127 &0.7291 &0.7304&0.7353 &0.7470 &0.7623 &0.7773 & 0.8019 &0.6160 &0.6370 &0.6397 &0.6384 \\
 SSDH & \underline{0.7646} & \underline{0.7761} & \underline{0.7931} &\underline{0.7958} &\underline{0.7638}& \underline{0.7735}&\underline{0.8037}&\underline{0.8182}&0.6408 &\underline{0.6691} & \underline{0.6811}& \underline{0.6817}\\
 \hline
 DOH &\textbf{0.8607} &\textbf{0.8739} &\textbf{0.8838} &\textbf{0.8863}&\textbf{0.8624} &\textbf{0.8686} &\textbf{0.8732} &\textbf{0.8702} &\textbf{0.7551} &  \textbf{0.7883}& \textbf{0.7916}& \textbf{0.7997}\\
 \hline
\end{tabular}
\end{center}
\end{table*}

\begin{figure*}[t]
\centering
\subfigure[MIRFlickr25k]{\includegraphics[width=0.3\textwidth,height=1.7in]{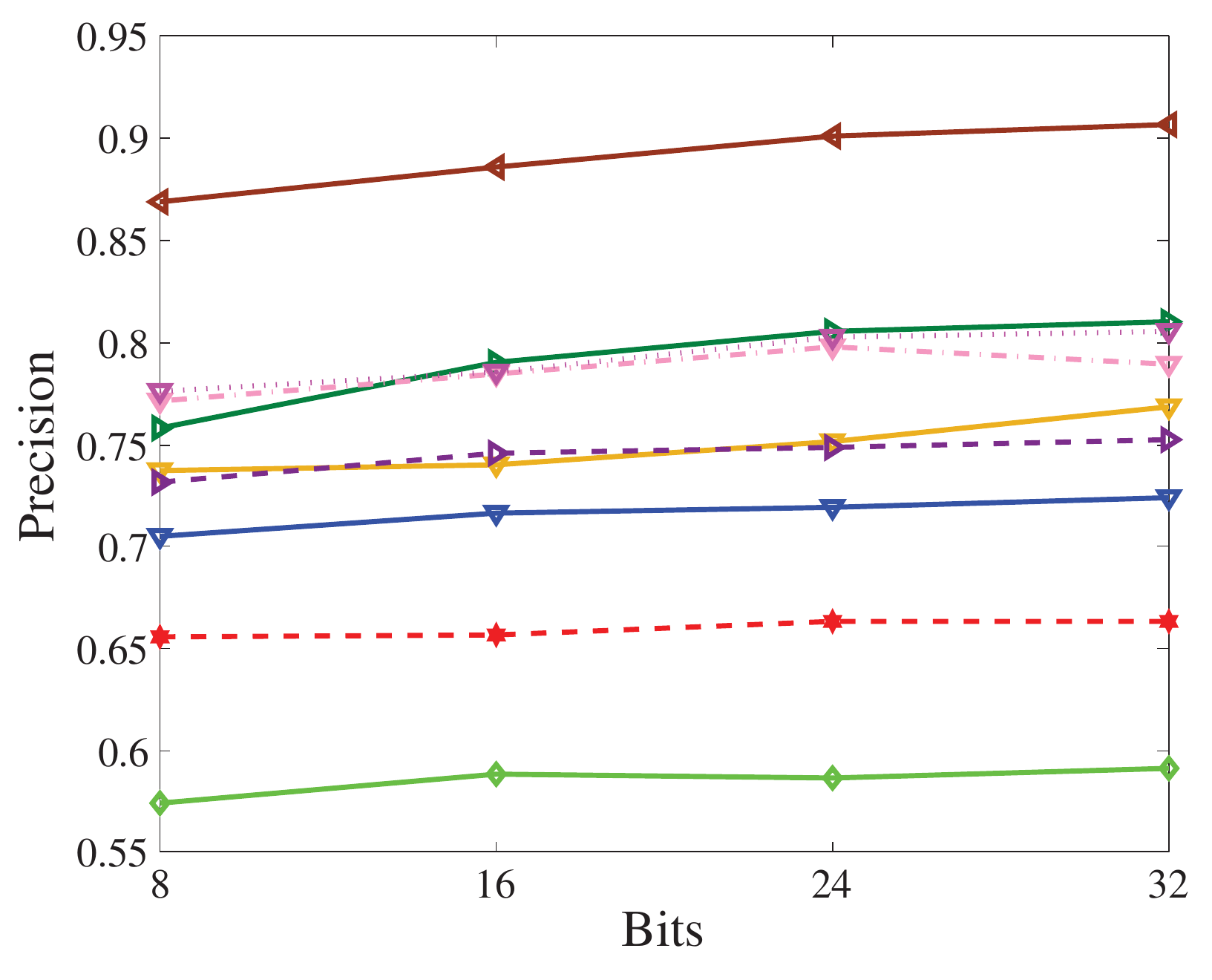}}
\subfigure[CIFAR-10]{\includegraphics[width=0.3\textwidth,height=1.7in]{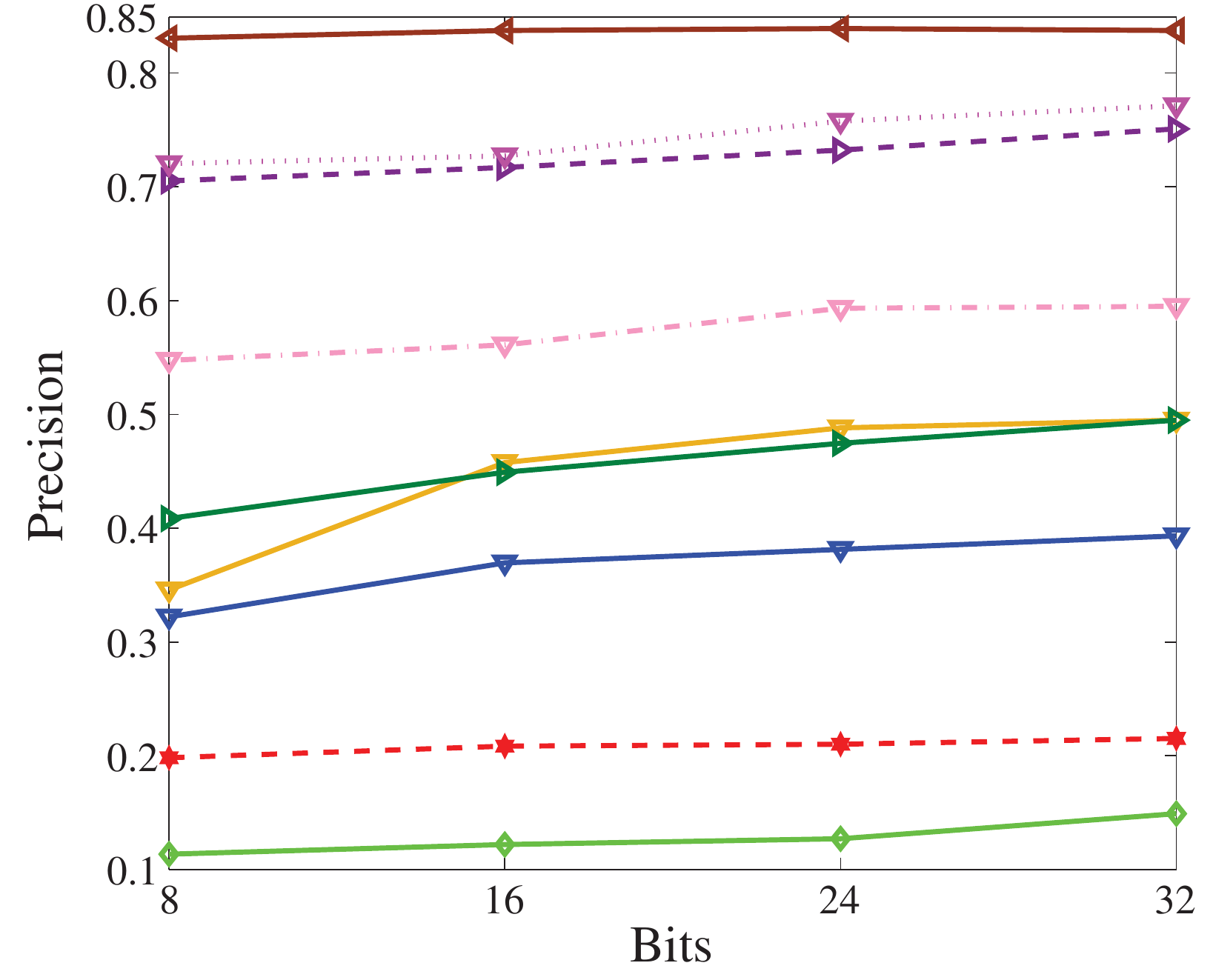}}
\subfigure[NUS-WIDE]{\includegraphics[width=0.38\textwidth,height=1.7in]{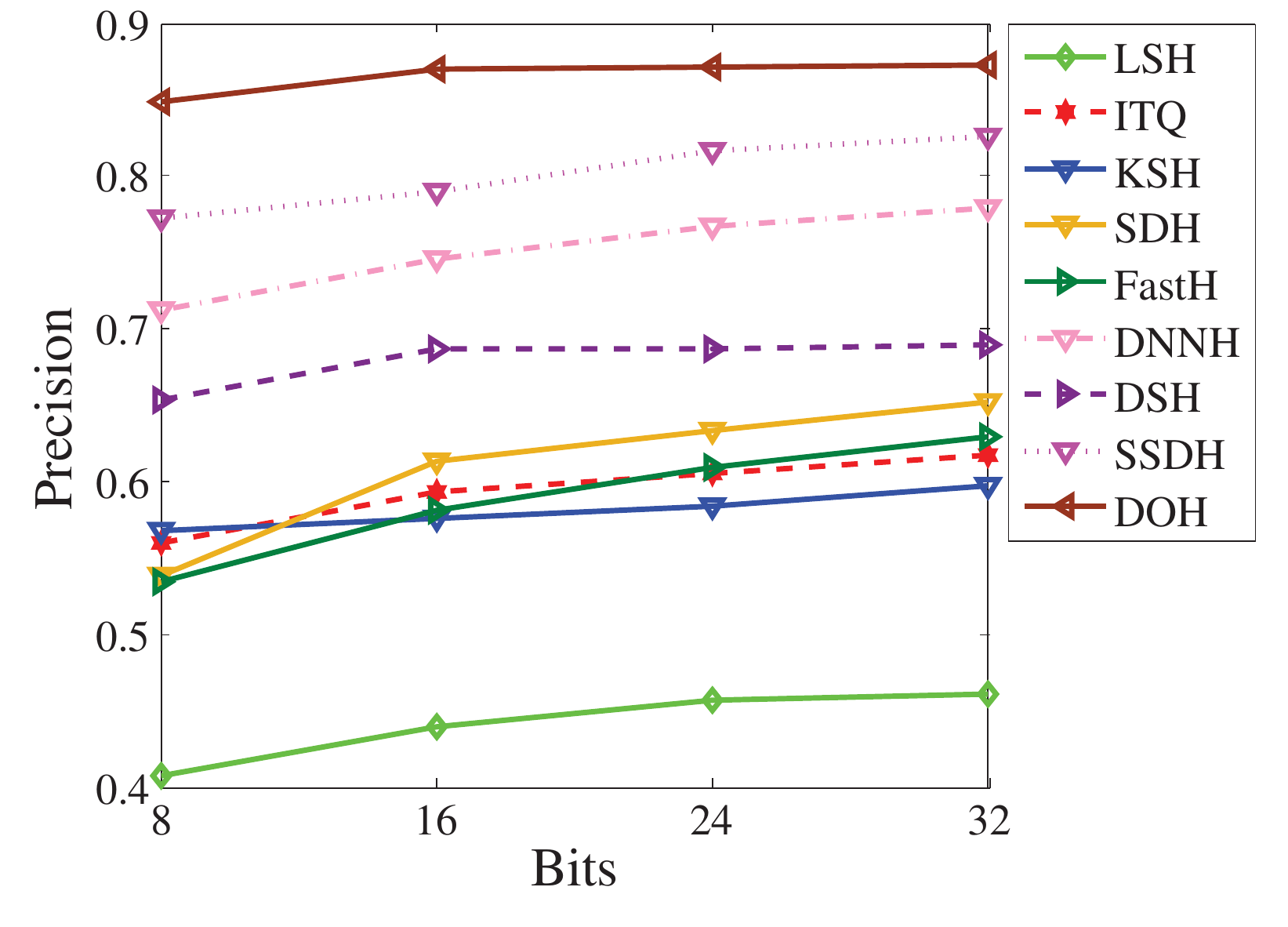}}
\caption{P@5000 curves of 16-bit hash code with respect to different code length on the three datasets.}
\label{Pre5000}
\end{figure*}

\subsection{Configurations of the networks}
DOH is a two-stream network where the FCN model aims to capture the local spatial information while the CNN model exploits the global semantic information. The detailed configurations of the FCN and CNN networks are described in Table~\ref{tab:network}. For the convolutional layers (e.g. $conv1$ to $conv5$), ``Filter" represents the number and the receptive filter size of the convolutional kernels as ``$number \times size \times size$", ``Pad" specifies the number of the pixels to add to the input, ``Stride" specifies the intervals at witch to apply the filter to the input, ``Pool" specifies the down-sampling operation, ``LRN" denotes the Local Response Normalization (LRN). For the fully connected layer, ``number" denotes the number of the outputs of this layer.

Additionally, ``ReLU" denoteS the Rectified Linear Unit activation function. ``Dropout" specifies whether the layer is regularized by dropout operation. More specifically, except for the $conv7$ layer adopting the global average pooling, the rest ``Pool"s adopt the max pooling strategy.
Additionally, DOH constructs a fully-connected layer (i.e., $fc$-$h$) to learn the ranking-based hash functions. Specifically, we set the number of the outputs as $K \times R$ for $fc$-$c$ layer where $K$ is the dimension of the feature space and $R$ is the bit length of hash codes. Different from the binary quantization hashing methods, the proposed DOH generates $K$-ary hash code. For fair comparison, we set $R = N_{c}/\log_{2}K$ when comparing with other binary quantization hashing methods.

\subsection{Experimental Settings}
We implement the proposed DOH method with the open-source Caffe \cite{jia2014caffe} framework on a NVIDIA K20 GPU server. We initialize the network with ``Xavier" initialization except for these layers including $conv1$ to $conv5$ and $fc6$ to $fc7$ that are copied from Alexnet pre-trained on ImageNet 2012 \cite{ILSVRC15}. As the remaining layers are trained from scratch, we set the learning rates as 100 times bigger than that of other layers for $fc$-$h$, as well as 10 times bigger for $conv6$, $conv7$, $fc8$ and $fc$-$c$. Our network is trained by using the mini-batch Stochastic Gradient Descent with the learning rate setting as $10^{-5}$. In all experiments, we fix the size of the min-batch as 64. As DOH involves one hyper-parameter, the dimension $K$ of the feature space, we use linear search in $\{2^{1},2^{2},2^{3},2^{4},2^{5}\}$ to select $K$. Specifically, we set $K$ as $2^2$ for MIRFlickr25K and Cifar10 datasets, and $2^3$ for NUS-WIDE dataset respectively.

\begin{figure*}[t]
\centering
\subfigure[MIRFlickr25k]{\includegraphics[width=0.3\textwidth,height=1.7in]{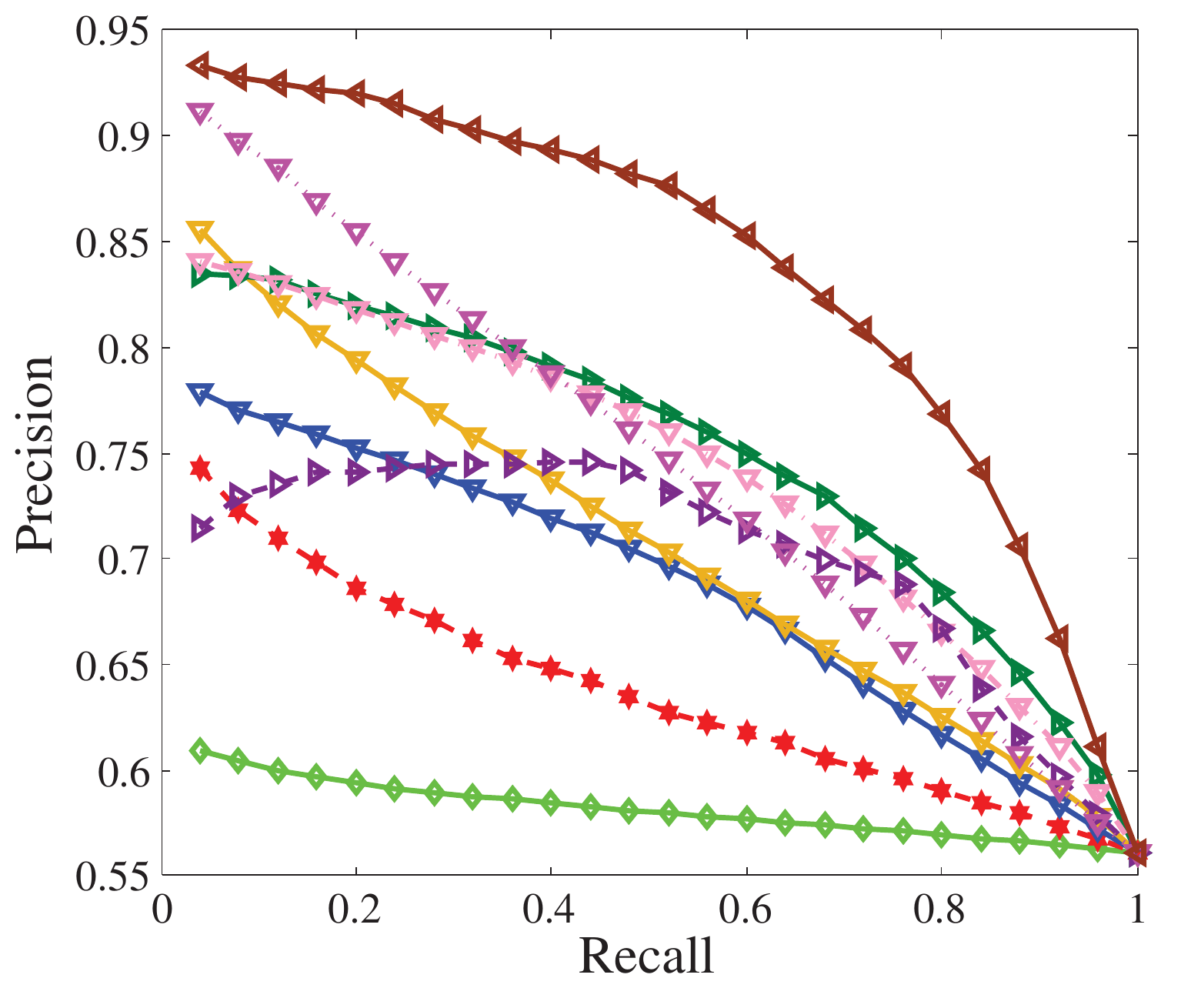}}
\subfigure[CIFAR-10]{\includegraphics[width=0.3\textwidth,height=1.7in]{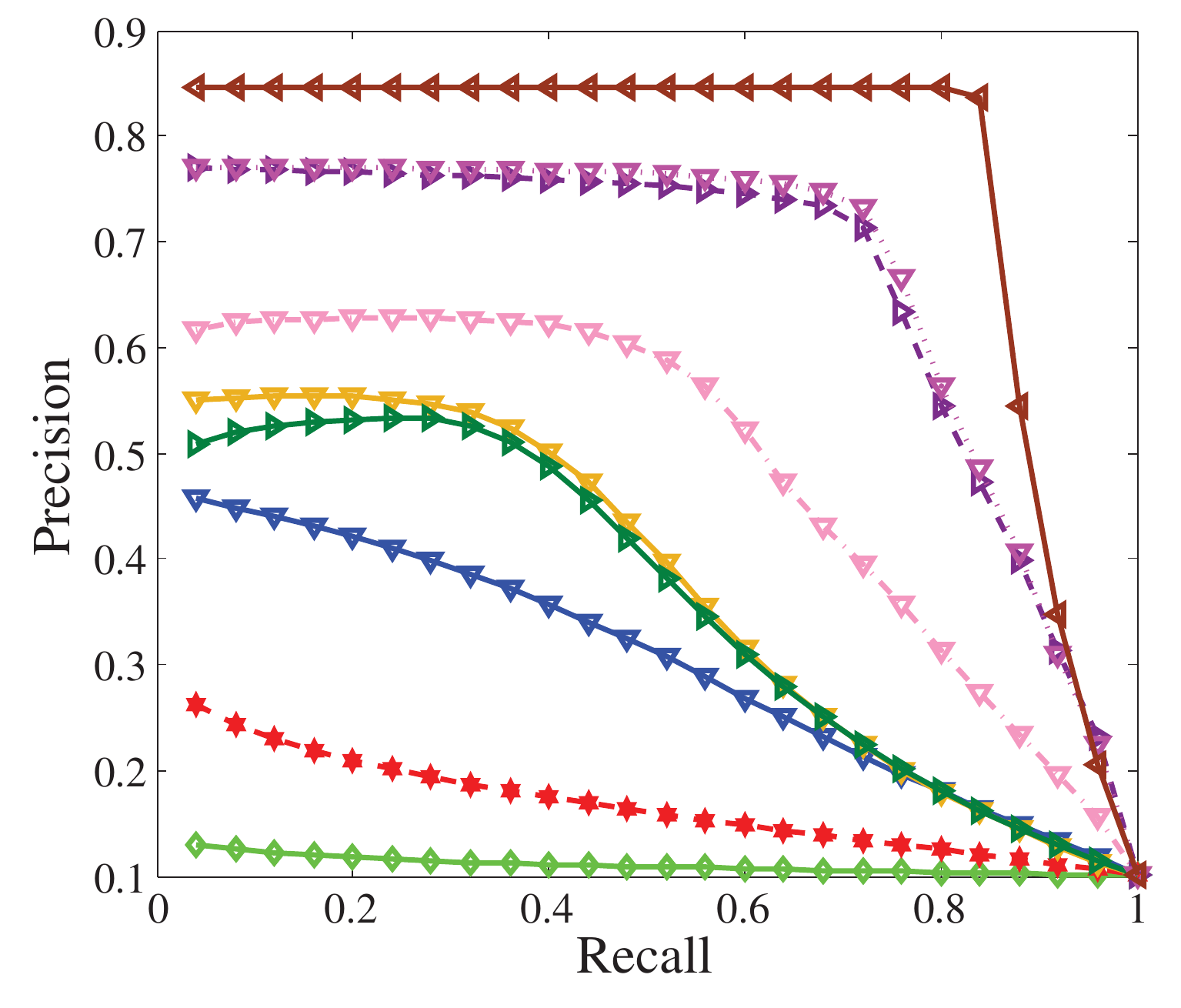}}
\subfigure[NUS-WIDE]{\includegraphics[width=0.38\textwidth,height=1.7in]{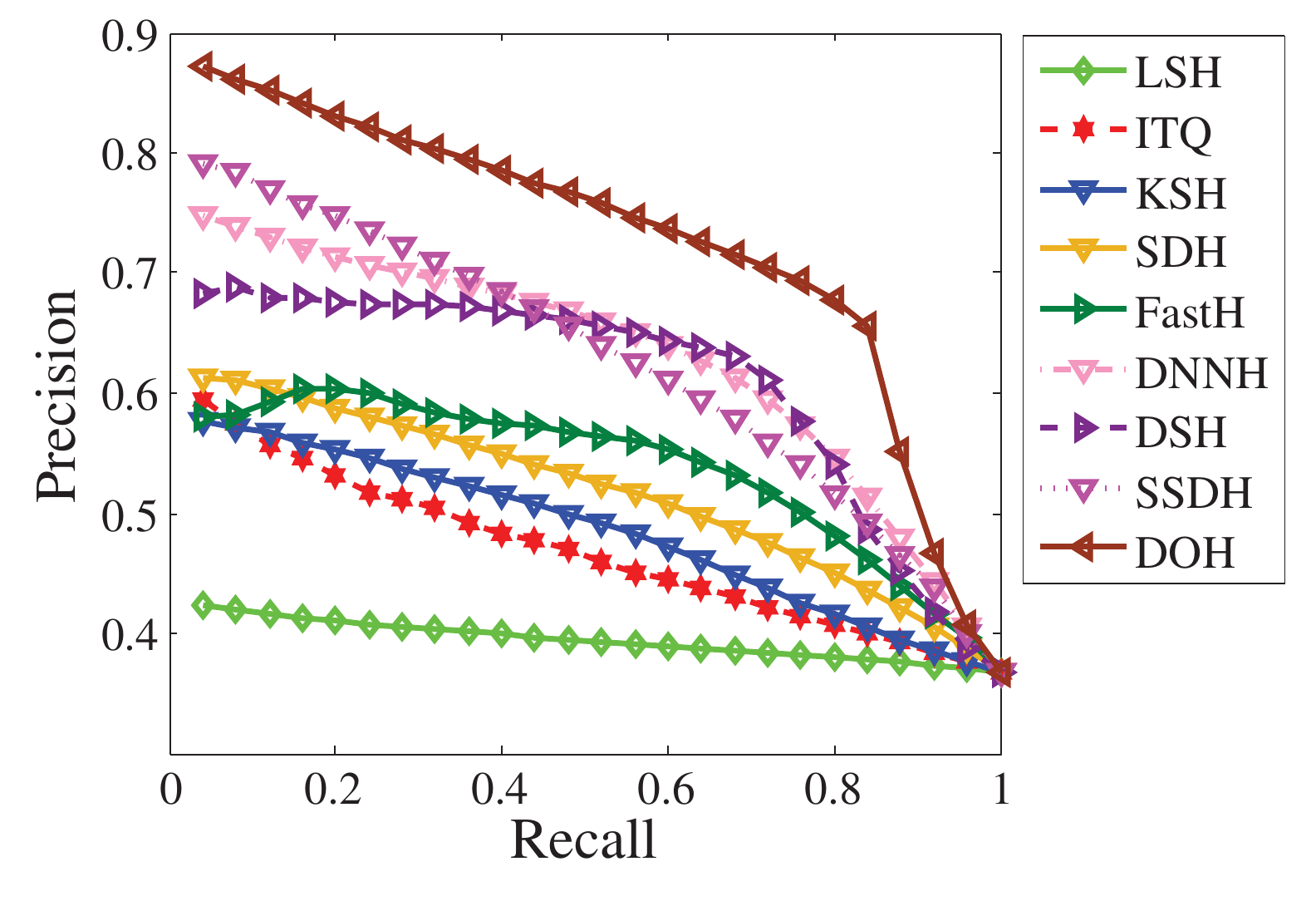}}
\caption{Precision-recall curves with respect to 16-bit hash code for different methods on the three datasets.}
\label{ROC}
\end{figure*}

\begin{figure*}[t]
\centering
\subfigure[MIRFlickr25k]{\includegraphics[width=0.3\textwidth,height=1.7in]{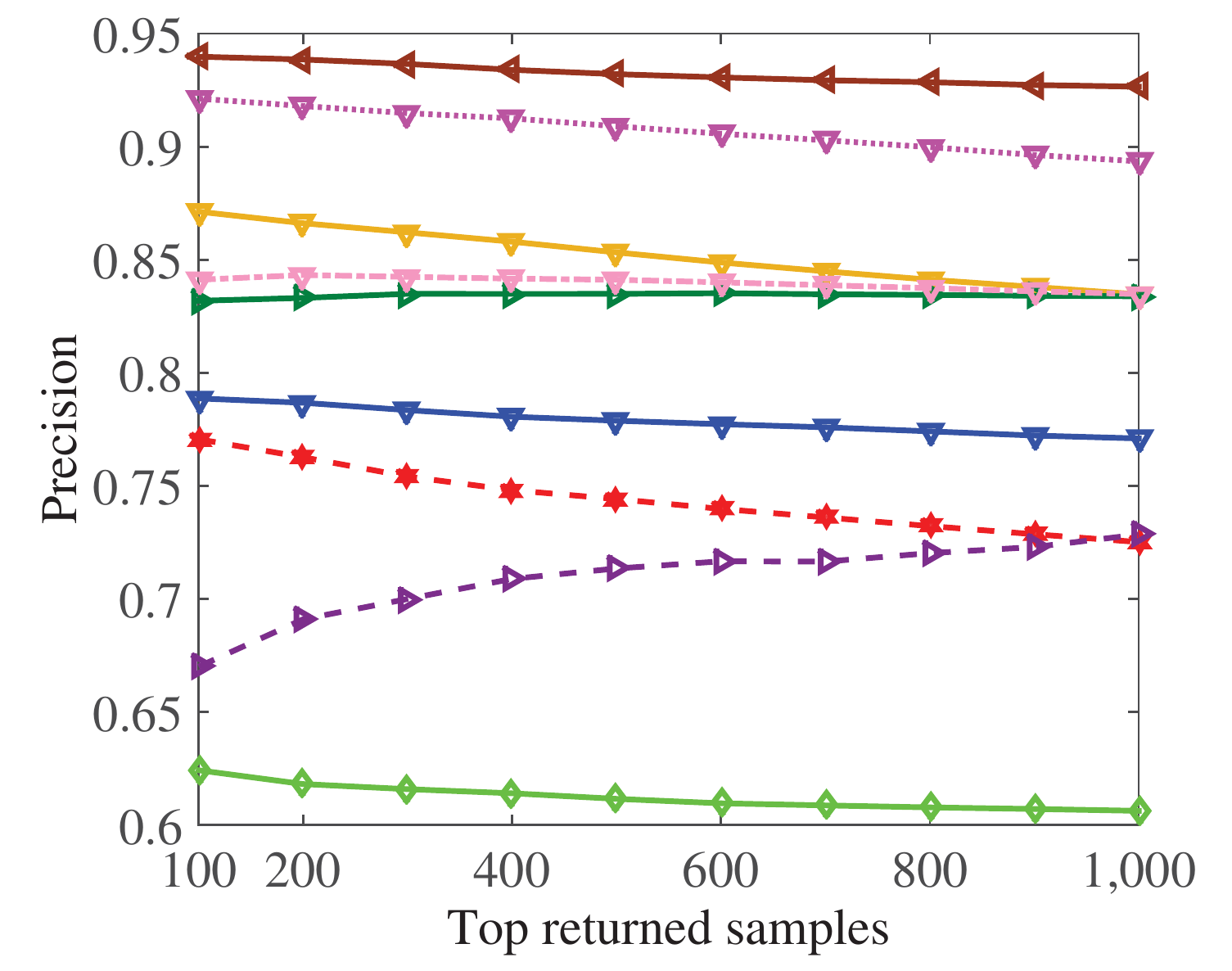}}
\subfigure[CIFAR-10]{\includegraphics[width=0.3\textwidth,height=1.7in]{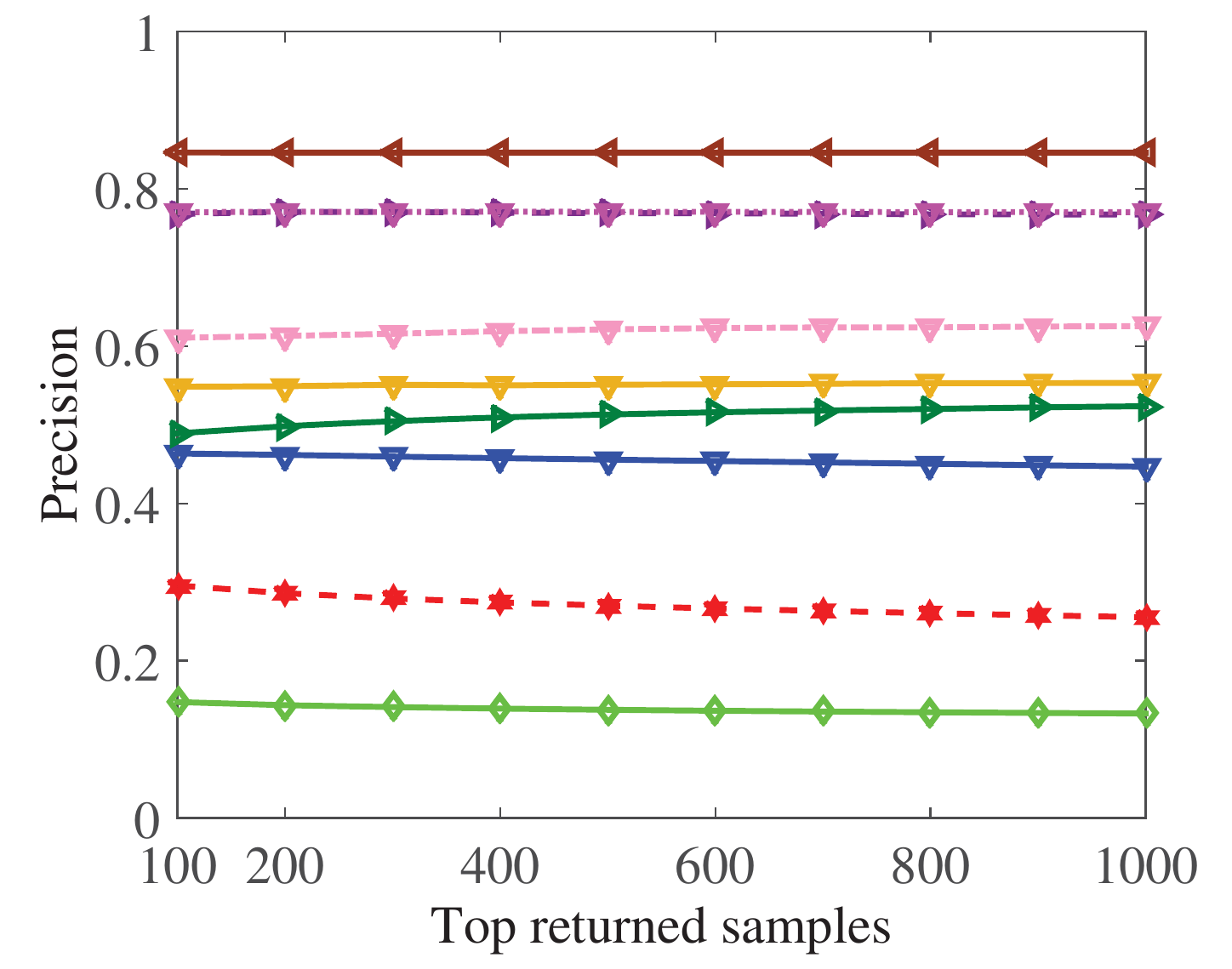}}
\subfigure[NUS-WIDE]{\includegraphics[width=0.38\textwidth,height=1.7in]{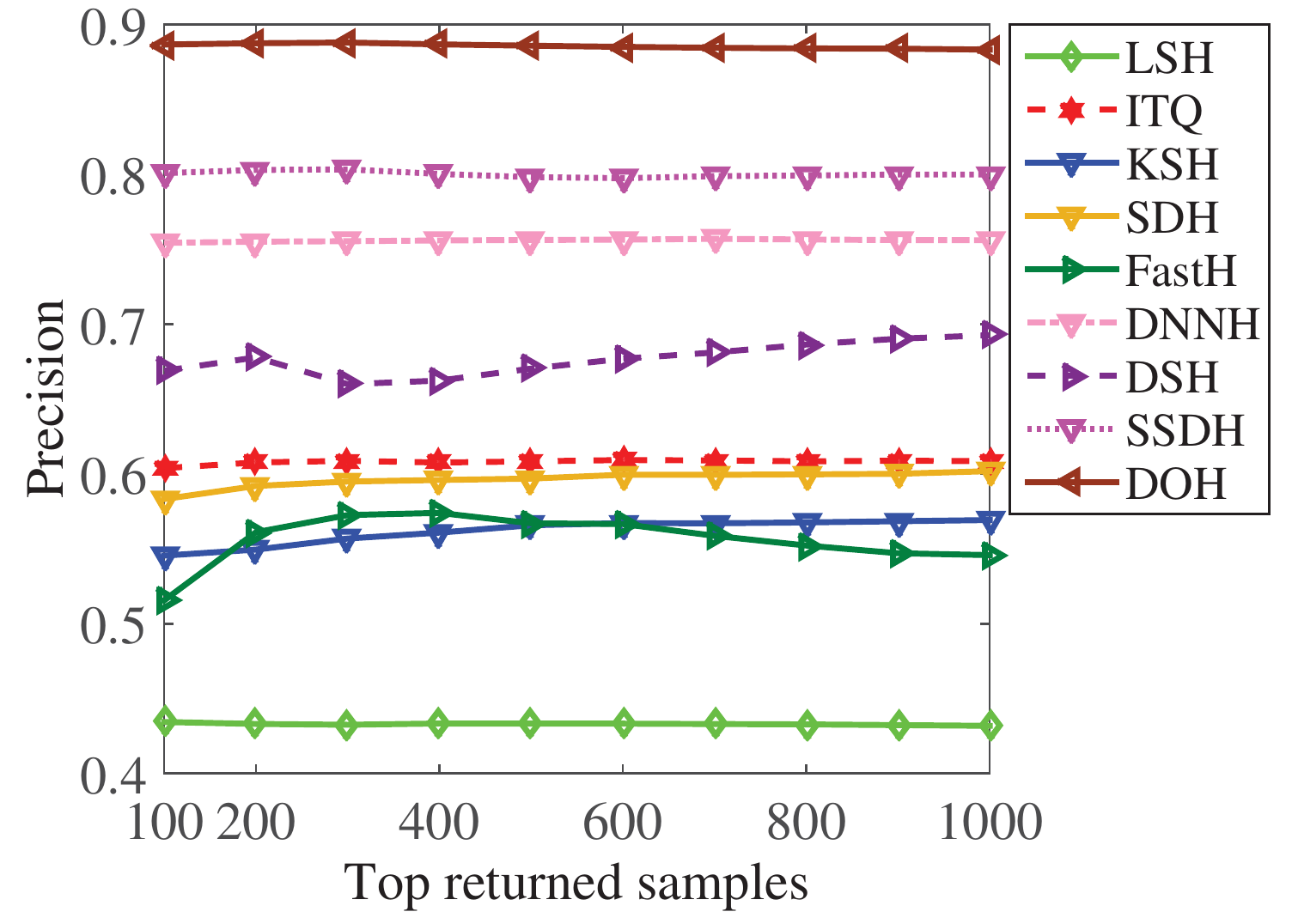}}
\caption{Precision curves of 16-bit hash code with respect to different number of top returned samples on the three datasets.}
\label{RrecisionCurves}
\end{figure*}

\begin{table*}
\begin{center}
\caption{mAP results of DOH and its variants with respect to different code length on three datasets. }
\label{tab:mAPV}
\begin{tabular}{ |c|cccc|cccc|cccc| }
\hline
  \multirow{2}{*}{Method}&
 \multicolumn{4}{c|}{MIRFLICKR25K} & \multicolumn{4}{c|}{CIFAR-10}& \multicolumn{4}{c|}{NUS-WIDE}\\
 \cline{2-13}
  & 8 bits & 16 bits & 24 bits&32 bits& 8 bits & 16 bits & 24 bits&32 bits& 8 bits & 16 bits & 24 bits&32 bits \\
 \hline
 \hline
  DOH-F &0.7980 &0.8325 & 0.8337 & 0.8385&0.8190 &0.8267 &0.8364 &0.8353 &0.7014 &0.7289 &0.7350 &0.7401 \\
 DOH-C  & 0.8193 &0.8627 &0.8735  &0.8783 &0.8452 &0.8585 & 0.8661&0.8669 &0.7382 &0.7701 &0.7793 &0.7870 \\
 DOH &\textbf{0.8607} &\textbf{0.8739} &\textbf{0.8838} &\textbf{0.8863}&\textbf{0.8624} &\textbf{0.8686} &\textbf{0.8732} &\textbf{0.8702} &\textbf{0.7551} &  \textbf{0.7883}& \textbf{0.7916}& \textbf{0.7997}\\
 \hline
\end{tabular}
\end{center}
\end{table*}

\subsection{Experimental Results}
\subsubsection{Comparison with the baselines}
We report mAP values for DOH and all the compared baselines in Tabel ~\ref{tab:map}. We can observe that DOH significantly outperforms all the compared baselines on different datasets with respect to different code length. In fact, compared to the best non-deep method (FastHash), DOH gains average performance increasements of $10.03{\rm{\% }}$, $37.56{\rm{\% }}$ and $22.13{\rm{\% }}$ for mAP values on MIRFlickr25K, Cifar10 and NUS-WIDE respectively. Furthermore, compared to SSDH, the best deep hashing method, DOH can still achieve average performance improvements of $9.42{\rm{\% }}$, $7.85{\rm{\% }}$ and $11.48{\rm{\% }}$ in terms of mAP values on the three datasets respectively. Such significantly improvements demonstrates the effectiveness of the proposed method.

In addition to mAP values, we also report the performance of P@5000 in terms of different code lengths and the precision curves of the 16-bit hash code with respect to different numbers of top returned samples in Figure~\ref{Pre5000} and Figure~\ref{RrecisionCurves}, respectively. When we compare DOH with FastH for the retrieval performance in terms of P@5000, the average performance gap between them are $13.24{\rm{\% }}$, $37.58{\rm{\% }}$ and $22.03{\rm{\% }}$ on the three datasets respectively. Similar, compared to SSDH, DOH grains $9.30{\rm{\% }}$, $7.87{\rm{\% }}$ and $11.37{\rm{\% }}$ performance improvements in average for P@5000 values on three datasets respectively. From Figure~\ref{RrecisionCurves}, we can observe that DOH also consistently outperforms all the compared methods on different datasets.


We plot the Precision-Recall (PR) curves of 16-bit hash code on three different datasets in Figure~\ref{ROC}. Specifically, the PR curve indicates the overall performance, and the larger value of the area under the PR curve reflects better performance. As can be seen in Figure~\ref{ROC}, DOH can achieve superior performance compared with the baselines on all the datasets. More specifically, DOH can yield higher precision at the lower recall points, which is satisfying for practical image retrieval system.

Overall, observed from the experimental results, DOH substantially outperforms all the compared baselines on different datasets in terms of mAP, P@5000 and PR curves. Such significant improvements verify the superiority of the proposed hashing method. We highlight three advantages of DOH in the following. Firstly, DOH can preserve the local discriminativity learned by the spatial attention model with FCN model to achieve effective image matching, while the other deep baselines only exploit the global semantic information with CNN model. Secondly, the ranking-based hash functions can produce discriminative hash codes by leveraging the relative ranking space structure. Finally, by jointly encoding the local spatial and global semantic information, DOH can capture useful ranking correlation structure to better preserve the ground truth similarities.

\begin{table*}
\begin{center}
\caption{P@5000 results of DOH and its variants with respect to different code length on three datasets.}
\label{tab:P5000V}
\begin{tabular}{ |c|cccc|cccc|cccc| }
\hline
  \multirow{2}{*}{Method}&
 \multicolumn{4}{c|}{MIRFLICKR25K} & \multicolumn{4}{c|}{CIFAR-10}& \multicolumn{4}{c|}{NUS-WIDE}\\
 \cline{2-13}
  & 8 bits & 16 bits & 24 bits&32 bits& 8 bits & 16 bits & 24 bits&32 bits& 8 bits & 16 bits & 24 bits&32 bits \\
 \hline
 \hline
  DOH-F & 0.8188& 0.8474& 0.8540 &0.8585 &0.7861 & 0.7821& 0.7925& 0.7888&0.7909 &0.8173 &0.8220 &0.8251 \\
 DOH-C  & 0.8401 & 0.8735& 0.8816 &0.8820 &0.8117 &0.8215 &0.8259 & 0.8276&0.8200 &0.8343 &0.8433 &0.8484 \\
 DOH &\textbf{0.8738} &\textbf{0.8852} &\textbf{0.8943} &\textbf{0.8959}&\textbf{0.8322} &\textbf{0.8355} &\textbf{0.8367} &\textbf{0.8325} &\textbf{0.8434} &  \textbf{0.8625}&\textbf{0.8641} &\textbf{0.8649}\\
 \hline
\end{tabular}
\end{center}
\end{table*}

\begin{figure*}[t]
\begin{center}
\includegraphics[width=1\textwidth]{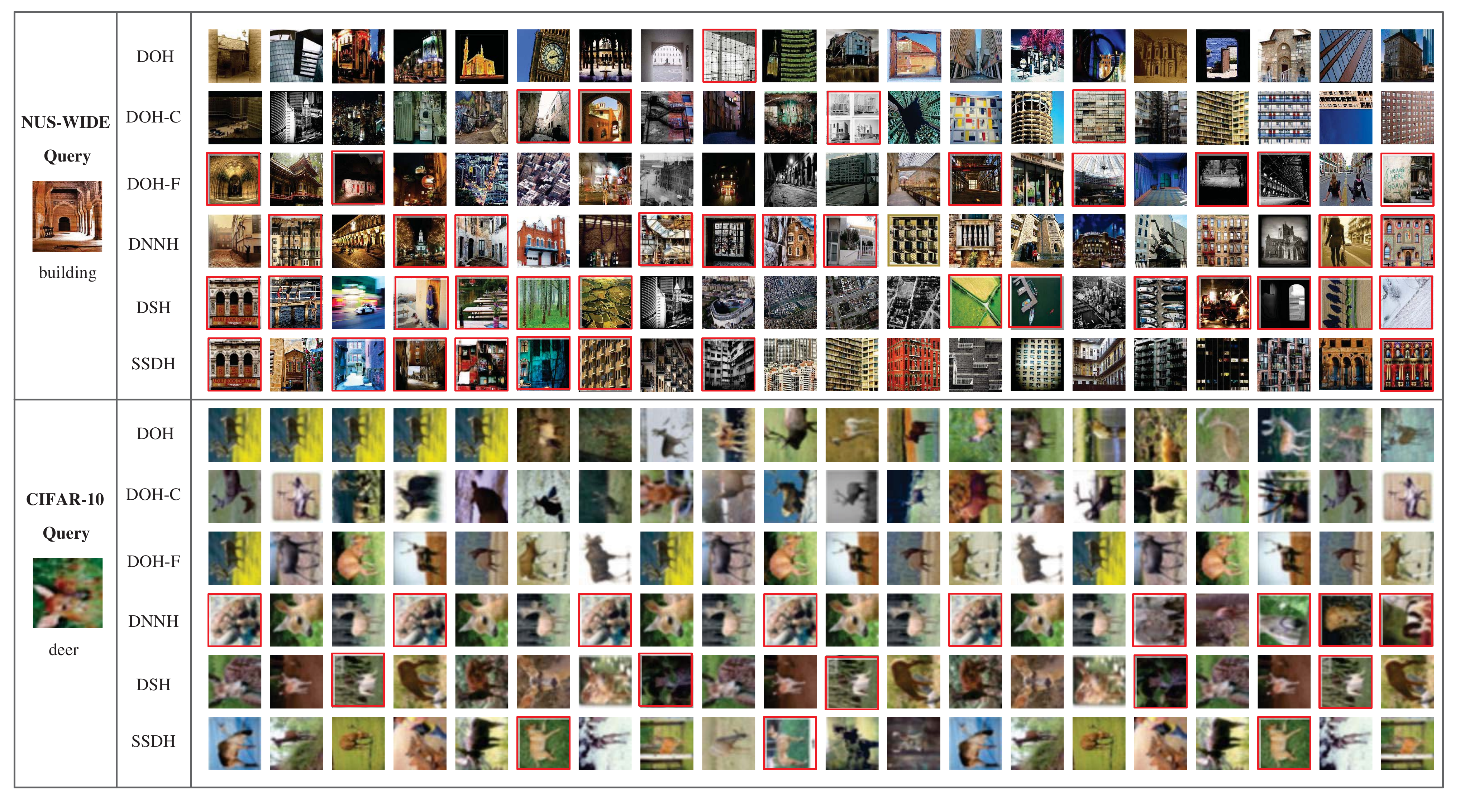}
\end{center}
\caption{Examples of retrieval results on the NUS-WIDE and CIFAR-10 datasets. The left column shows the query images. The middle column shows different methods, and their corresponding retrieval results are shown in the right column. The code length of all the methods is set as 32 bits and the top 20 results obtained using Hamming ranking are shown. The red rectangles indicate the wrong retrieval results. }
\label{fig:retrieval_samples}
\end{figure*}

\subsubsection{Comparison with two variants}
In the proposed deep network, we construct a subnetwork followed after both of FCN and CNN to learn the unified ordinal representations $\mathbf{h}$ which are utilized to approximate the ranking-based hash functions. A possible alternative to this subnetwork is that two independent fully-connected layers are adopted to learn different ordinal representations for FCN and CNN respectively. Therefore, the ordinal representation learned in this manner can only gain the knowledge from either the local spatial information or the global semantic information. Specifically, we investigate two variants of DOH: (1) \textbf{DOH-F}, variant only using the FCN network to learn the ordinal representation that is local aware; (2) \textbf{DOH-C}, variant only using the CNN metwork to learn the ordinal representation that is global aware.

The retrieval performance in terms of mAP is illustrated in Table \ref{tab:mAPV}. We find that, by exploiting the local spatial and global semantic information simultaneously, DOH can consistently outperforms both two variants on the three datasets. For example, compared to DOH-F, DOH can gain average performance improvements of $5.05{\rm{\% }}$, $3.93{\rm{\% }}$ and $5.73{\rm{\% }}$ on MIRFlickr25K, CIFAR10 and NUS-WIDE, respectively. Compared to DOH-C, the average performance gap is $1.77{\rm{\% }}$, $0.95{\rm{\% }}$ and $1.50{\rm{\% }}$ on the three datasets.

In addition to mAP, the performance of P@5000 is shown in Table \ref{tab:P5000V}. Overall, the proposed method substantially outperforms the two variants. In detail, compared to DOH-F, DOH achieves the average performance improvements of $4.26{\rm{\% }}$, $4.69{\rm{\% }}$ and $4.49{\rm{\% }}$ for the three datasets. Similarly, compared to DOH-C, DOH gains the average performance improvements of $1.80{\rm{\% }}$, $1.26{\rm{\% }}$ and $2.22{\rm{\% }}$ on  MIRFlickr25K, CIFAR10 and NUS-WIDE, respectively.

Another interesting finding is that the retrieval performance of DOH and its two variants in terms of mAP and P@5000 substantially outperforms SSDH, the best deep hashing method that adopts the binary quantization function. 
Actually, DOH and its two variants are similar in the sense that they all involve to exploit rank correlation spaces with deep networks. Therefore, the ranking structure leveraged by the proposed hashing method is useful to generate discriminative hash code for yielding superior retrieval performance.

We also show some retrieval results of the top 20 returned samples in terms of Hamming ranking on the NUS-WIDE and CIFAR-10 dataset in Figure~\ref{fig:retrieval_samples}. We can observe that DOH and its two variants can achieve the best retrieval results than the deep baselines, which indicates the effectiveness of exploiting the ranking structure with the deep network. Specifically, we note that DOH can yield better retrieval results than DOH-F and DOH-C on NUS-WIDE dataset. This indicates that the ranking-based hash function learned by jointly exploiting the local spatial and global semantic information can better preserve the ground truth similarity of the image pair.



\begin{figure}
\centering
\subfigure[mAP]{\includegraphics[width=0.23\textwidth]{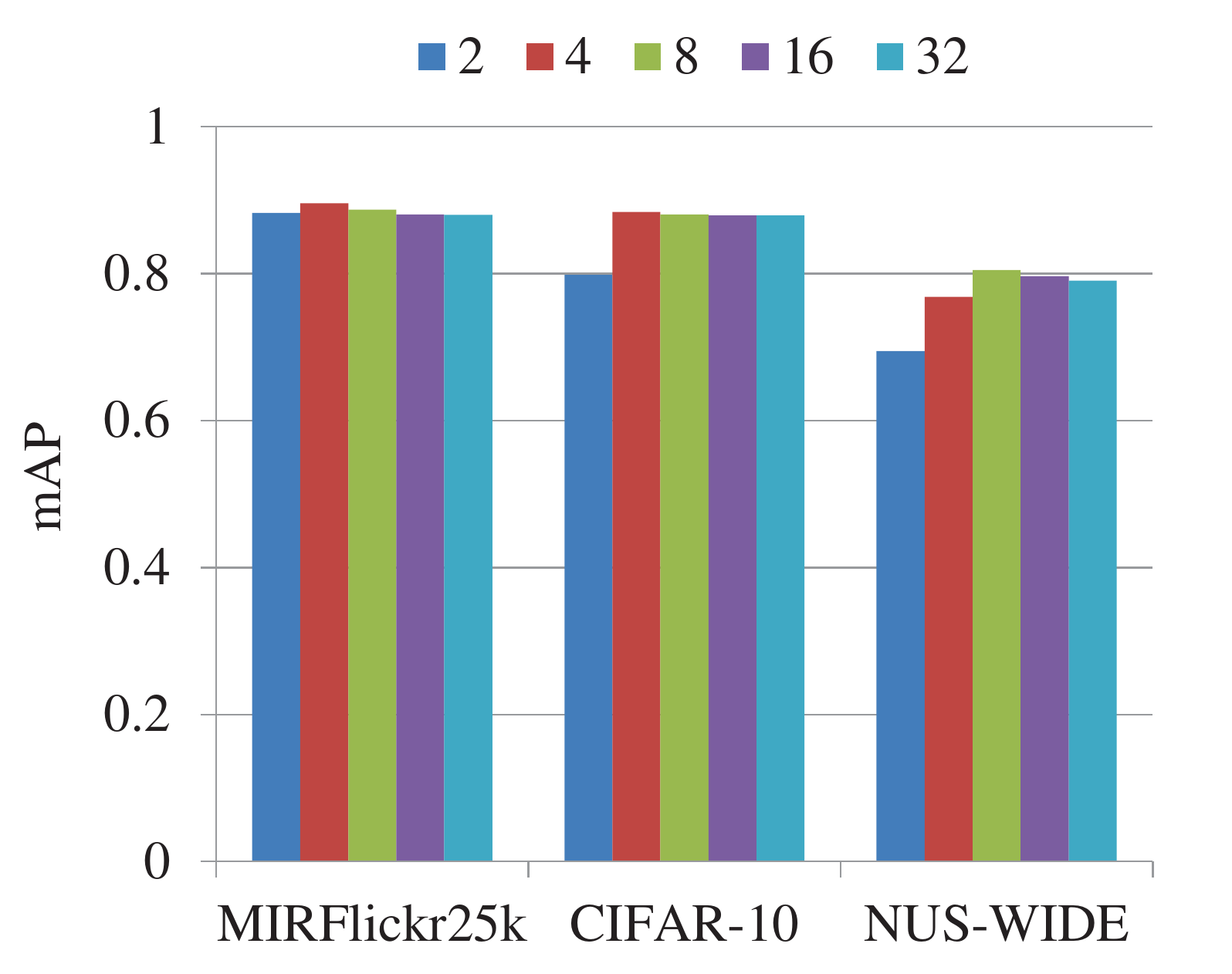}}
\subfigure[P@5000]{\includegraphics[width=0.23\textwidth]{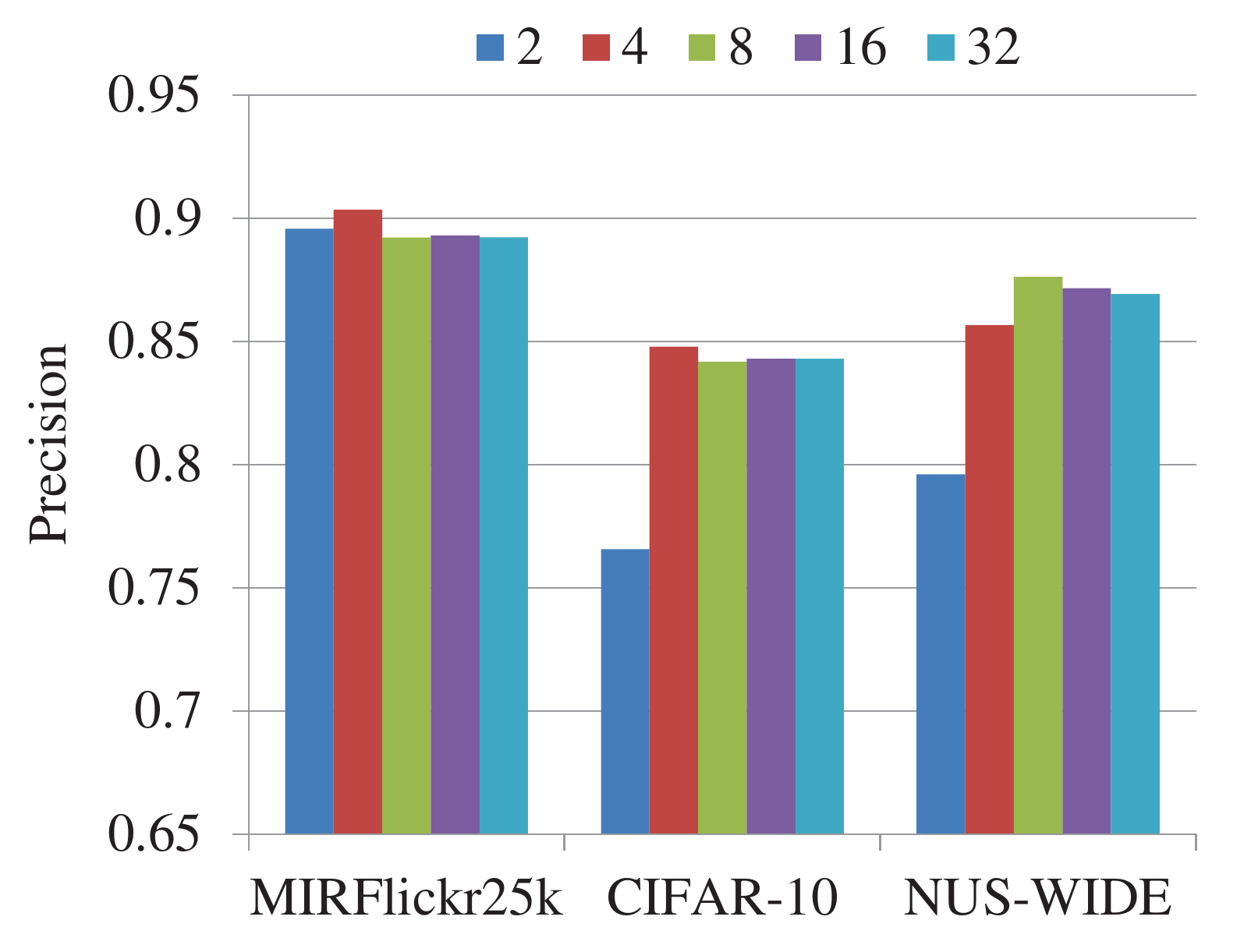}}
\caption{The mAP and P@5000 with respect to different $K$ for 60-bits hash code on the three datasets.}
\label{Para}
\end{figure}

\subsubsection{Effect of parameter}
The proposed DOH involves one parameter, the dimension $K$ of the feature space. In order to verify the sensitivity, we conduct experiments to analyze the influence on different datasets by using linear search in $\{2^{1},2^{2},2^{3},2^{4},2^{5}\}$. Specifically, we set the code length as 60 as it is the least common multiplier of $\log_{2}K$. The retrieval performance in terms of mAP and P@5000 are shown in Figure \ref{Para}. For MIRFlickr25k dataset, there is very small performance influence under different settings of $K$. However on CIFAR-10 and NUS-WIDE, the performance slightly decreases when setting $K$ as 2. As illustrated in Figure \ref{Para}, we can set $K$ as 4 for MIRFlickr25k and CIFAR-10, and $K$ as 8 for NUS-WIDE, respectively.


\section{Conclusion}
In this work, we propose a novel deep hashing (DOH) method to learn the ranking-based hash functions by exploiting the rank correlation space from both the local and global views. Specifically, a two-stream network is designed to learn the unified ordinal representation for approximating the ranking-based hashing functions by exploiting the local spatial information from the FCN network and the global semantic information from the CNN network simultaneously. More specifically, for the FCN network, an effective spatial attention model is proposed to capture the local discriminativity by learning well-specified locations closely related to target objects. By jointly leveraging such local information learned with the spatial attention model and the global semantic information, DOH can lean high quality ordinal representation to produce discriminative hash codes, thus achieving superior performance for image retrieval. 
Extensive experimental results on three datasets verify the superiority of the proposed DOH method in learning discriminative hash codes for image retrieval.

\bibliographystyle{ieeetr}
\bibliography{sigproc}

\ifCLASSOPTIONcaptionsoff
  \newpage
\fi

\end{document}